 \documentclass{amsart}
 \usepackage{amsfonts,amssymb}

 \usepackage{amsfonts,amsmath,amssymb,cite,graphicx,color,ulem,amsthm}
 \usepackage{fourier}
 \usepackage[margin=1.5in]{geometry}

 \theoremstyle{plain}
 \newtheorem{theorem}{Theorem}[section]
 \newtheorem{proposition}[theorem]{Proposition}
 \newtheorem{lemma}[theorem]{Lemma}
 \newtheorem{assumption}[theorem]{Assumption}

 \newtheorem{remark}[theorem]{Remark}

\def\thetab{\boldsymbol{\theta}}

\def\etab{\boldsymbol{\eta}}

\def\ab{\boldsymbol{a}}
\def\bb{\boldsymbol{b}}
\def\Bb{\boldsymbol{B}}

\def\hb{\boldsymbol{h}}

\def\Rb{\boldsymbol{R}}
\def\xb{\boldsymbol{x}}
\def\vb{\boldsymbol{v}}

\def\zb{\boldsymbol{z}}
\def\Zb{\boldsymbol{Z}}
\def\Wb{\boldsymbol{W}}
\def\Thetab{\boldsymbol{\Theta}}

\def\d{\displaystyle}

\def\RR{\mathbb{R}} \def\NN{\mathbb{N}} 
\def\EE{\mathbb{E}}\def\PP{\mathbb{P}}\def\SS{\mathbb{S}}

\DeclareMathOperator{\supp}{supp}

\DeclareMathOperator{\argmin}{argmin}
\DeclareMathOperator{\cov}{cov}

\def\<{\langle} \def\>{\rangle}

\begin{document}

\title{Trainability and Accuracy of Neural Networks:\\ An Interacting Particle System Approach}
% \title{Training Accuracy of Neural Networks:\\ An Interacting Particle System Approach}
% \title{Accuracy of Neural Networks After Training:\\ An Interacting Particle System Approach}
\author{Grant M. Rotskoff}
\author{Eric Vanden-Eijnden}
 \address{Courant Institute of Mathematical Sciences, New York
   University, 251 Mercer street, New York, NY 10012}
 \email{rotskoff@cims.nyu.edu, eve2@cims.nyu.edu} 
\thanks{We thank Joan Bruna and
  Weinan E for discussions about the approximation error of neural
  networks, and Sylvia Serfaty for her insights about interacting
  particle systems. We are also grateful to  Andrea Montanari, Matthieu Wyart, and the anonymous referee for their helpful comments.}

\begin{abstract}
  Neural networks, a central tool in machine learning, have
  demonstrated remarkable, high fidelity performance on image
  recognition and classification tasks.  These successes evince an
  ability to accurately represent high dimensional functions, but
%   potentially of great use in computational and applied mathematics.
%   That said,
% Networks, however, require to be optimized or `trained' and
   rigorous results about the approximation
  error  of neural networks after training are few. Here we establish conditions for global convergence of the standard optimization algorithm used in
  machine learning applications, stochastic gradient descent (SGD), and quantify the scaling of its error with the size of the network. This is done by reinterpreting SGD as the
  evolution of a particle system with interactions governed by a
  potential related to the objective or ``loss'' function used to
  train the network. We show that, when the number $n$ of units
  is large, the empirical distribution of the particles descends on a
  convex landscape towards the global minimum at a rate independent of $n$, with a resulting approximation error that universally scales as
  $O(n^{-1})$. These properties are established in the form of a Law of Large Numbers and a Central Limit Theorem for
  the empirical distribution.
%   and, remarkably, these scaling results do not depend on the
%   dimensionality of the domain of the function that we seek to
%   represent.  
Our analysis also quantifies the scale and nature of the
  noise introduced by SGD and provides
  guidelines for the step size and batch size to use when training a
  neural network. We illustrate our findings on examples in which we
  train neural networks to learn the energy function of the continuous
  3-spin model on the sphere.  The approximation error scales as our
  analysis predicts in as high a dimension as $d=25$.
\end{abstract}

\maketitle

\tableofcontents

\section{Introduction}
\label{sec:intro}

While both speech recognition and image classification remain active areas of research, extraordinary progress has been made on both problems---ones that appeared intractable only a decade
ago~\cite{LeCun:2015dt}. By harvesting the power of neural networks
while simultaneously benefiting from advances in computational
hardware, complex tasks such as automatic language translation are now
routinely performed by computers with a high degree of
reliability. The underlying explanation for these significant advances
seems to be related to the expressive power of neural networks, and
their ability to accurately represent high dimensional functions.

These successes open exciting possibilities in applied and
computational mathematics % among many other areas of science
that are 
only beginning to be explored~\cite{Behler:2007fe ,Schneider:2017dn,
  Khoo:2018wz, Berg:2017tg,E:2017a,E2017b,Zhang:2018}. Any numerical
calculation that uses a given function begins with a
finite-dimensional approximation of that function. Because standard
approximations, e.g., Galerkin truncations or finite element
decompositions, suffer from the curse of dimensionality, it is nearly
impossible to scale such methods to large
dimensions~$d$. Fundamentally, these representations are linear
combinations of basis functions. The issue arises because the
dimensionality of the representation is equal to that of the
truncation. Neural networks, on the other hand, are highly nonlinear
in their adjusting parameters. As a result, the effective
dimensionality of a neural network is much higher than its total number of parameters,
which may explain the impressive function approximation capabilities observed in practice, even when $d$ is large. Characterizing this
observation with analysis is non-trivial though, precisely because the
representation of a function by a neural network is nonlinear in its
parameters. This renders many of the standard tools of numerical
analysis useless, since they are in large part based on linear
algebra.

The significant achievements of machine learning have inspired many
efforts to provide theoretical justification to a vast and growing
body of empirical knowledge. At the core of our understanding of
the approximation properties of neural networks are the well-known ``Universal Approximation Theorems'' that
specify the conditions under which a neural network can represent a
target function with arbitrary
accuracy~\cite{Cybenko:1989fm,Barron:1993ba,Park:2008ka}. 
Despite the power of these results, they do not indicate how the network parameters should be
optimized to achieve maximal accuracy in practice\cite{Bottou:2003wp}. In particular, these theorems do not
provide general guidance on how the error of the network scales with its size at the end of training.
Several recent papers have focused on the analysis of the
shape and properties of the objective or ``loss'' function
landscape~\cite{Sagun:2014tg, Choromanska:2014ui,Baitsy:2018}.  These
studies have mainly focused on the fine features of this landscape,
trying to understand how non-convex it is and making analogies with
glassy landscapes.  Additionally, some analysis has been performed in
cases where the number of parameters vastly exceeds the amount of
training data, a setting that guarantees convexity and dramatically
simplifies the landscape.  Further studies have examined the dynamics
of the parameters on the loss landscape to understand the properties
of optimization procedures based on stochastic gradient descent.

In this paper, we adopt a different perspective which enables powerful
tools for analysis.  Similar to what was recently
proposed in~\cite{Mei:2018,Sirignano:2018vg,Chizat:2018}, we view the parameters
in the network as particles and the loss function as a potential that
dictates the interaction between them. Correspondingly, training the
network can be interpreted as the evolution of the particles in this
interaction potential.  Using the interchangeability of the $n$ interacting particles / parameters in the neural representation, we focus on their empirical distribution  and analyze its properties
 when  $n$ is large using standard limit
theorems\cite{kipnis2013scaling,Serfaty:2015,leble2017large,Serfaty:2017kb}.
This viewpoint allows us to bypass many of the difficulties that arise
with approaches that attempt to study the dynamics of the individual
particles. In particular:

\begin{enumerate}
\item We derive an evolution equation for the empirical distribution of the
  particles, and show that it evolves by gradient descent in the 2-Wasserstein metric on a convex energy landscape. This observation allows us to assert that convergence towards equilibrium of the empirical
  distribution occurs on a time scale that is independent of $n$ to
  leading order---similar results were obtained
  in~\cite{Mei:2018,Sirignano:2018vg,Chizat:2018}. The results are
  obtained in the form of Law of Large Numbers (LLN) for the empirical
  distribution of the parameters. As a consequence, we rederive the
  Universal Approximation Theorem and establish that it can be realized dynamically.
\item We  quantify the fluctuations of the empirical distribution at finite $n$ above its limit. We show that these fluctuations are of order $O(n^{-1/2})$ and controlled at all $t<\infty$. In addition, we establish conditions under which these fluctuations heal and become $O(n^{-1})$ as $t\to\infty$. These results rely on a Central Limit Theorem (CLT) and indicate that the neural network approximation error is universal and scales as $O(n^{-1})$ as $n\to\infty$ in any $d$.
\end{enumerate}
\noindent
These results are established first in situations where gradient
descent (GD) on the loss function is used to optimize or ``train'' the parameters in the
network, and then shown to also apply in the context of stochastic
gradient descent (SGD). In the latter case, our analysis sheds light
on the nature of the noise introduced in SGD, and indicates how the
time step and the batch size should be scaled to achieve the optimal
error. We briefly elaborate on these statements below, first precisely formulating the problem.

\subsection{Problem set-up}
\label{sec:setup}

Given a function $f:\Omega\to\RR$ defined on the closed manifold
$\Omega \subseteq \RR^d$, consider its approximation by a neural network of the form
\begin{equation}
  \label{eq:21}
  f^{(n)}(\xb) = \frac1n \sum_{i=1}^n c_i \hat \varphi(\xb,\zb_i)
\end{equation}
where $n\in \NN$, $(c_i,\zb_i) \in D \equiv \RR \times \hat D $ are
parameters to be learned for $i=1,\ldots, n$, and
$\varphi: \Omega \times D \to \RR$ is some function---we assume
throughout this paper that $\hat D$ is a closed manifold in
$\RR^N$. The function $\hat \varphi$ is usually referred to as the
`nonlinearity' or `unit' and $n$ as the width of the network. To
simplify notations, we use
$\thetab = (c_i,\zb)\in D$ and
$\varphi(\xb,\thetab) = c\hat \varphi(\xb,\zb)$, in terms of
which~\eqref{eq:21} reads
\begin{equation}
  \label{eq:21compact}
  f^{(n)}(\xb) = \frac1n \sum_{i=1}^n \varphi(\xb,\thetab_i)
\end{equation}
Many models used in machine learning can be cast in
the form~\eqref{eq:21}-\eqref{eq:21compact}:

\begin{itemize}
\item \textbf{Radial basis function networks.} In this case
  $\hat D \equiv\Omega$ and $\hat\varphi( \xb,\zb) \equiv \phi(\xb-\zb)$
  where $\phi$ is some kernel, for example that of a radial function
  such as
  \begin{displaymath}
    \phi(\xb) = \exp\left(-\tfrac12 \kappa |\xb|^2\right)
  \end{displaymath}
  where $\kappa>0$ is a fixed constant.
\item \textbf{Single hidden layer neural networks.} In this case,
  $\hat D \subset \SS^{d}$ and $\hat \varphi( \xb,\zb)= \hat \varphi(\xb,\ab,b)$
  with e.g. $\ab\in \SS^{d-1}$, $b\in[-1,1]$, and
  \begin{displaymath}
    \varphi(\xb,\ab,b) = h(\ab\cdot \xb+b)
  \end{displaymath}
  where $h:\RR\to\RR$ is e.g. a sigmoid function $h(z) = 1/(1+e^{-z})$
  or a rectified linear unit (ReLU) $h(z) = \max(z,0)$.
\item \textbf{Iterated neural networks.} These are structurally similar
  to single hidden layer neural networks. For example, to construct
  a two-layer network we take $h$ as above and for $m\in \NN$,
  $m\le d$ define $\hb^{(1)}:\RR^m \to \RR^m$ such that
  \begin{displaymath}
    h_j^{(1)}(\vb) = h(v_j), \qquad \vb =(v_1,\ldots, v_m)\in \RR^m, \quad j=1,\ldots, m
  \end{displaymath}
then set
\begin{displaymath}
  f^{(n)}(\xb) = \frac1n \sum_{i=1}^n c_i h\left(\ab_i^{(0)} \cdot
  \hb^{(1)}\left(A_i^{(1)} \xb + \bb_i^{(1)}\right)+b_i^{(0)}\right)
\end{displaymath}
where $\ab_i^{(0)} \in \RR^m$, $b_i^{(0)}\in\RR$,
$A_i^{(1)} \in \RR^{m\times d}$, $\bb_i^{(1)}\in \RR^m$,
$i=1,\ldots,n$. Therefore here we have
$\zb = (\ab^{(0)},b^{(0)},A^{(1)} ,\bb^{(1)})\in \hat D\subset
\RR^{m+1+m\times d + m}$ (where with a slight abuse of notation we
view the matrix $A^{(1)}$ has a vector in $\RR^{m\times
  d}$). Three-layer networks, etc. can be constructed similarly.  Note
that our results apply to deep neural networks when their final layer
grows large, with fixed depth.
\end{itemize}
%
% \begin{enumerate}
% \item How good can the approximation~\eqref{eq:21}-\eqref{eq:21compact} be if we optimize
%   $\{c_i,\zb_i\}_{i=1}^n = \{\thetab_i\}_{i=1}^n$?
% \item Can we guarantee the convergence of the commonly used
%   optimization algorithms?
% \end{enumerate}
%
To measure the discrepancy between the target function $f$ and its neural network approximation $f^{(n)}$, we need to introduce a distance, or loss
function, between~$f$ and~$f^{(n)}$. A natural candidate often used in
practice is
\begin{equation}
  \label{eq:21a}
  \mathcal{L}[f^{(n)}] = \tfrac12\int_{\Omega} \big|f(\xb) -
   f^{(n)}(\xb)\big|^2 \nu(d\xb) = \tfrac12 \EE_\nu \big|f -
   f^{(n)}\big|^2
\end{equation}
where $\nu$, the  data distribution, is some positive
measure on $\Omega$ such that $\nu(\Omega)<\infty$ (for example the
Hausdorff measure on $\Omega$, which we will denote by $d\xb$).  We can view $\mathcal{L}[f^{(n)}]$ as an objective
function for $\{\thetab\}_{i=1}^n$:
\begin{equation}
  \label{eq:51}
  \mathcal{L}[f^{(n)}] = C_f
  -\frac1n\sum_{i=1}^n  F(\thetab_i) + \frac1{2n^2}
  \sum_{i,j=1}^n  K(\thetab_i,\thetab_j)
\end{equation}
where $C_f=\frac12\EE_\nu\left|f\right|^2$ and
we defined
\begin{equation}
  \label{eq:18}
  F(\thetab) = c \hat F(\zb), \qquad K(\thetab,\thetab') = cc' \hat K(\zb,\zb')
\end{equation}
with
\begin{equation}
  \label{eq:22a}
  \begin{aligned}
    \hat F(\zb) &= \EE_\nu [f \hat\varphi(\cdot,\zb)],
    \\
    \hat K(\zb,\zb') & = \EE_\nu [\hat\varphi(\cdot,\zb)
    \hat\varphi(\cdot,\zb')] \equiv \hat K(\zb',\zb).
  \end{aligned}
\end{equation}
Trying to minimize~\eqref{eq:51} over $\{\thetab_i\}_{i=1}^n$ leads
to difficulties, however, since this is potentially (and presumably) a
non-convex optimization problem, which typically has local
minimizers. In particular, if we perform training by making
$\{\thetab_i\}_{i=1}^n$  evolve via gradient descent (GD) over the
loss, i.e. if we use
\begin{equation}
  \label{eq:34aa}
      \dot\thetab_{\!i} = \nabla F(\thetab_{\!i}) -
      \frac1n\sum_{j=1}^n \nabla K(\thetab_{\!i},\thetab_{\!j}),
\end{equation}
there is no guarantee \textit{a~priori} that these parameters will
reach the global minimum of the loss or even a local minimum with a
value for the loss that is close to that of the global minimum. As a
result determining the value of~\eqref{eq:21a} after training (and its
scaling with $n$, say) is nontrivial. It is therefore
natural to ask:
\begin{quote}
   {\it  How accurate is the approximation~\eqref{eq:21}-\eqref{eq:21compact}  if we optimize
  $\{\thetab_i\}_{i=1}^n$ by applying the algorithms commonly used in machine learning?}
\end{quote} 
This is the main question
we investigate in the present paper.

\subsection{Main results and organization}
We will consider the evolution in time  of the representation
\begin{equation}
  \label{eq:68aa}
  f^{(n)}_t = \frac1n \sum_{i=1}^n \varphi(\cdot,\thetab_i(t))
\end{equation}
and study the behavior of this function for large $n$ and large
$t$. To this end we use tools from interacting particle systems, and
also build on known results about the nature of the loss
function~\eqref{eq:51} in the limit as $n\to\infty$---these results
are recalled in Sec.~\ref{sec:continuousintro}.

\medskip

In Sec.~\ref{sec:weighted} we consider the situation where
$\thetab_i(t)$ are the solution of the GD flow~\eqref{eq:34aa}---as
explained below, this is somewhat of an idealized situation since we
typically must work with the empirical loss rather than the exact one,
but it is more easily amenable to analysis. By looking at the
evolution of the empirical distribution of $\{\thetab_i(t)\}_{i=1}^n$
rather than that of the parameters themselves, we establish a Law of
Large Numbers (LLN) for $f^{(n)}_t$ namely that
$\lim_{n\to\infty} f^{(n)}_t= f_t$, where $f_t$ evolves as
\begin{equation}
  \label{eq:25}
  \partial_t f_t(\xb) = -\int_\Omega M_t(\xb,\xb') \left(f_t(\xb') -
    f(\xb')\right) \nu(d\xb').
\end{equation}
where $f$ is the target function and $M_t(\xb,\xb')$ a positive
semi-definite kernel whose form is explicit---see Proposition~\ref{th:llnft}.
The evolution equation~\eqref{eq:25} can be interpreted as GD for $f_t$ over the
loss in some metric inherited from the 2-Wasserstein metric, and in Proposition~\ref{th:lln} we show the flow converges to the target function, i.e., 
\begin{equation}
  \label{eq:82}
  \lim_{t\to\infty} f_t =\lim_{t\to\infty} \lim_{n\to\infty} f^{(n)}_t = f. 
\end{equation}
We also establish that the limit in $n$ and $t$ commute. 
Regarding the scaling of the fluctuations above $f_t$ when $n$ is
finite, in  Proposition~\ref{th:cltg} we  establish a Central Limit Theorem (CLT) that asserts that
these fluctuations are of size $O(n^{-1/2})$, i.e.
$n^{1/2} (f^{(n)}_t - f_t)$ has a limit in law as $n\to\infty$. In
addition, in Proposition~\ref{th:cltglt} we show that these fluctuations are controlled at all times, and in Proposition~\ref{th:cltgvlt} that under certain conditions they  heal as $t\to\infty$, in the sense that
\begin{equation}
  \label{eq:55zT}
  f^{(n)}_{a_n} = f + O(n^{-1}) \qquad \text{as} \quad
  n\to\infty \quad \text{with} \quad a_n /\log n
  \to\infty.
\end{equation}

\medskip

In Sec.~\ref{sec:stochgrad} we analyze the typical situation in which it is not possible to calculate~\eqref{eq:21a} or~\eqref{eq:22a} exactly. Rather, we must approximate these
expectations using a ``training set'', i.e. a set of points
$\{\xb_p\}_{p=1}^P$ distributed according to~$\nu$ on which $f$ is known, so that instead of
$\mathcal{L}[f^{(n)}]$ we must use
\begin{equation}
  \label{eq:57}
  \mathcal{L}_P[f^{(n)}] = \frac1P \sum_{p=1}^P |f(\xb_p)-f^{(n)}(\xb_p)|^2
\end{equation}
and instead of $\hat F$ and $\hat K$
\begin{equation} \label{eq:58} \hat F_P(\zb) = \frac1P \sum_{p=1}^P
  f(\xb_p) \hat \varphi(\xb_p,\zb), \quad \hat K_P(\zb,\zb') =
  \frac1P \sum_{p=1}^P \hat \varphi(\xb_p,\zb)
  \hat\varphi(\xb_p,\zb').
\end{equation}
If in~\eqref{eq:34aa} we replace $F(\thetab)$ and
$K(\thetab,\thetab')$ by their empirical estimates over a subset of the training points
$F_P(\thetab) = c \hat F_P(\zb)$ and
$K_P(\thetab,\thetab') = cc' \hat K_P(\zb,\zb')$, we arrive at what is referred
to as stochastic gradient descent (SGD)---the method of
choice to train neural networks. 
We focus on situations in which we
can redraw the training set as often as we need, namely, at every step
during the learning process, an algorithm called online learning. 
In this
case, in the limit as the optimization time step $\Delta t$ used in SGD
tends to zero, SGD becomes asymptotically equivalent to an SDE whose
drift terms coincide with those of GD but with multiplicative noise
terms added. In this context, we establish that~\eqref{eq:25}
and~\eqref{eq:82} also hold if we choose the size $P$ of the batch used
in~\eqref{eq:58} at every SGD step such that as $P=O(n^{2\alpha})$
with $\alpha>0$. Regarding the scaling of the fluctuations,  if we set
$\alpha\in (0,1)$, we lose accuracy
and~\eqref{eq:55zT} is replaced by
\begin{equation}
  \label{eq:55zTSGD}
  f^{(n)}_{a_n} = f + O(n^{-\alpha}) \qquad \text{as} \quad
  n\to\infty \quad \text{with} \quad a_n /\log n
  \to\infty.
\end{equation}
However if $\alpha\ge1$, we get \eqref{eq:55zT} back (meaning also
that there is no advantage to take $\alpha$ bigger than 1). These results are stated in Propositions~\ref{th:train} and \ref{th:train2}.

In Sec.~\ref{sec:examples} we illustrate these results, using
a spherical $p$-spin model with $p=3$ as test 
function to represent with a neural network. We show that the network
accurately approximates this function in up to $d=25$ dimensions, with
a scaling of the error consistent with the results established in
Secs.~\ref{sec:weighted} and~\ref{sec:stochgrad}. These results are
obtained using both a radial basis function network, and a
single hidden layer network using sigmoid functions.

Concluding remarks are made in Sec.~\ref{sec:conclu} and in
an Appendix we establish a finite-temperature
variant (Langevin dynamics) of~\eqref{eq:55zT} which applies when
additive noise terms are added in the GD
equations~\eqref{eq:34aa}. This result reads
\begin{equation}
  \label{eq:55}
  \lim_{T\to-\infty} f^{(n)}_t = f + n^{-1} \tilde f_t + o(n^{-1})  \quad \text{with} \quad
  \tilde f_t = \beta^{-1} \epsilon^* + \beta^{-1/2}\tilde \epsilon_t
\end{equation}
where $T$ is the time at which we initiate the training. Here
$\beta> 0$ is a parameter playing the role of inverse temperature,
$\epsilon^*:\Omega\to\RR$ is some given (non-random) function and
$\tilde \epsilon_t :\Omega\to\RR$ is a Gaussian process with mean
zero and covariance
$\EE [\tilde \epsilon_t(\xb) \tilde\epsilon_t(\xb')]\propto
\delta(\xb-\xb') $. Note that \eqref{eq:55} gives~\eqref{eq:55zT} back
after quenching (i.e. by sending $\beta\to\infty$).  The result
in~\eqref{eq:55} is stated in Proposition~\ref{th:cltfT}

\bigskip

As we have emphasized, our approach has strong ties with the statistical
mechanics of systems of large numbers of interacting particles. 
Our main aim here
is to introduce a framework showing how results and concepts developed
in this context are useful to address
questions in machine learning.
Conversely, we seek to illustrate that ML provides new mathematical questions about an interesting class of particle systems. With this in mind, we adopt a presentation style that relies on formal asymptotic arguments to
derive our results, though we are confident that providing rigorous proofs
to our propositions is achievable. To a certain extent, this program
was already started in~\cite{Mei:2018,Chizat:2018, Sirignano:2018vg}.

%\section{Preliminaries: Functional formulation}
\section{Functional formulation of the learning problem}
\label{sec:continuousintro}

As discussed in Bach~\cite{Bach:2017}, it is useful to give conditions
under which~\eqref{eq:21} has a limit as $n\to\infty$, for two main
reasons: First it shows which functions can be represented as
in~\eqref{eq:21} if we allow the number of units $n$ to grow to
infinity, and second, while the loss
function~\eqref{eq:21a} may be non-convex for $\{\thetab_i\}_{i=1}^n$, the limiting functional for the parameter distribution is convex.

\subsection{Universal Approximation Theorem}

Consider the space $\mathcal{F}_1$ of all functions that can be represented as 
\begin{equation}
  \label{eq:52}
  f = \int_{\hat D} \hat \varphi(\cdot,\zb) \gamma(d\zb)
\end{equation}
where $\gamma$ is some (signed) Radon measure on $D$ with finite total
variation ($L^1$-norm),
$|\gamma|_{\text{TV}}=\int_{\hat D} |\gamma(d\zb)|<\infty$: we will
denote the space of these Radon measures by $\mathcal{M}(\hat D)$ and
that of probability measures by $\mathcal{M}_+(\hat D)$. The space $\mathcal{F}_1$ is important in our context, since any $f\in \mathcal{F}_1$ can be realized  as
\begin{equation}
  \label{eq:116}
  f = \lim_{n\to\infty} \frac1n \sum_{i=1}^n c_i\hat \varphi(\cdot, \zb_j)
\end{equation}
by drawing $\{c_i,\zb_i\}_{i\in \NN}$ as follows.  Start from the
Jordan decomposition for $\gamma$~\cite{billing},
\begin{equation}
  \label{eq:17}
  \gamma = \gamma_+-\gamma_-,
\end{equation}
where $\gamma_+$ and $\gamma_-$ are positive measures with
$\supp \gamma_+ \cup \supp\gamma_- = \supp \gamma$ and
$\supp \gamma_+ \cap \supp\gamma_- = \emptyset$. Using this
decomposition, we can  draw $\zb_i$'s independently from
$(\gamma_++\gamma_-)/|\gamma|_{\text{TV}}\in\mathcal{M}_+(\hat
D)$, where
$|\gamma|_{\text{TV}} = \int_{\hat
  D}(\gamma_+(d\zb)+\gamma_-(d\zb)) < \infty$ , and set $c_i=+|\gamma|_{\text{TV}}$
if $\zb_i\in \supp \gamma_+$ and $c_i=-|\gamma|_{\text{TV}}$ if
$\zb_i\in \supp \gamma_-$. By the Law of Large Numbers we then have
\begin{equation}
  \gamma_n = \frac1n \sum_{i=1}^n c_i\delta_{\zb_i}\rightharpoonup
  \gamma \qquad \text{as} \quad n\to\infty\label{eq:28}
\end{equation}
which implies~\eqref{eq:116}. We can also use the Central Limit
Theorem to get an approximation error at finite~$n$, a calculation we carry out in Sec.~\ref{sec:clt}.

Since the space $\mathcal{F}_1$ depends on the choice of unit $\hat\varphi$, to characterize it we make:
\begin{assumption}
  \label{th:ascomp} Both the input space $\Omega$ and the feature space $\hat D$ are closed (i.e. compact with no boundaries) smooth Riemannian manifolds. The unit is continuously differentiable in $\zb$, i.e. $\forall \xb\in \Omega$, $\hat\varphi(\xb,\cdot) \in C^1(\hat D)$.
\end{assumption}
\begin{assumption}[Discriminating unit]
  \label{th:as1} The unit satisfies
  \begin{equation}
    \label{eq:50}
    \int_{\Omega} g(\xb) \hat \varphi(\xb,\cdot) \nu(d\xb) = 0 \quad
    \text{a.e.\ \ in \ \ $\hat D$} \quad
    \Rightarrow \quad g = 0 \quad \text{a.e.\ \ in\ \ $\Omega$}
  \end{equation}
\end{assumption}
\noindent
The differentiability of $\hat\varphi$ in $\zb$ is required to guarantee uniqueness of the GD flow.
% Under these assumptions we have:
%
\begin{theorem}[Universal Approximation Theorem~\cite{Cybenko:1989fm,Barron:1993ba, Park:2008ka}]
  \label{th:1}
  Under Assumptions~\ref{th:ascomp} and~\ref{th:as1}, $\mathcal{F}_1$ is a dense subspace of $L^2(\Omega,\nu)$, i.e. given any
  $f\in L^2(\Omega,\nu)$ and $\epsilon >0$, there exists
  $\gamma^* \in \mathcal{M}(\hat D)$
  such that $|\gamma^*|_{\text{TV}}<\infty $ and 
  \begin{equation}
  \label{eq:29}
  f^* = \int_{\hat D} \hat \varphi(\cdot, \zb) \gamma^*(d\zb) \in \mathcal{F}_1
\end{equation}
satisfies
  \begin{equation}
    \label{eq:123}
    \| f- f^*\|_{L^2(\Omega,\nu) } \le \epsilon.
  \end{equation}
\end{theorem}
\noindent A similar theorem was originally stated in~\cite{Cybenko:1989fm}. Since its proof is elementary let us reproduce it here: 
\begin{proof}The space $\mathcal{F}_1$ is a linear subspace of $L^2(\Omega, \nu)$ since, if $f=\int_{\hat D} \hat \varphi(\cdot,\zb) \gamma(d\zb) \in \mathcal{F}_1$,
  \begin{displaymath}
    \begin{aligned}\|f\|^2_{L^2(\Omega,\nu)} &= \int_\Omega \left(\int_{\hat D} \hat \varphi(\xb,\zb) \gamma(d\zb)\right)^2\nu(d\xb) \\
    &= \int_{\hat D \times \hat D} \hat K(\zb,\zb') \gamma(d\zb)\gamma(d\zb')\\
    & \le \|\hat K\|_\infty |\gamma |^2_{\text{TV}} < \infty
    \end{aligned}
  \end{displaymath}
  where we used $\|\hat K\|_\infty = \sup_{(\zb,\zb')\in \hat D \times \hat D} |\hat K(\zb,\zb')|<\infty$ which follows from Assumption~\ref{th:ascomp}.
  To show that $\mathcal{F}_1$ is dense in $L^2(\Omega, \nu)$, we proceed by contradiction. Assuming that $\mathcal{F}_1$ is not dense, by the Hahn-Banach theorem, there exists a nonzero linear functional $L:L^2(\Omega, \nu)\to\RR$ such that $Lf=0$ for all $f\in\mathcal{F}_1$. By the Riesz representation theorem, the action of $L$ on $f$ can be represented as the inner product in $L^2(\Omega, \nu)$ between $f$ and some $g\in L^2(\Omega, \nu)$, i.e.  there must exist $g\not=0$ such that for all $f=\int_{\hat D} \hat \varphi(\cdot,\zb) \gamma(d\zb)\in \mathcal{F}_1$ (i.e. all $\gamma \in \mathcal{M}(\hat D)$ with finite variation)
  \begin{displaymath}
    \begin{aligned}0 &= \int_\Omega g(\xb) \left(\int_{\hat D} \hat \varphi(\xb,\zb) \gamma(d\zb)\right)\nu(d\xb) \\
    &= \int_{\hat D} \left( \int_\Omega g(\xb) \hat \varphi(\xb,\zb) \nu(d\xb) \right) \gamma(d\zb),
    \end{aligned}
  \end{displaymath}
  This requires that 
  \begin{displaymath}
    0=\int_\Omega g(\xb) \hat \varphi(\xb,\cdot ) \nu(d\xb)\quad \text{a.e.\ \ in \ \ $\hat D$}. 
  \end{displaymath}
  which, by Assumption~\ref{th:as1}, implies that $g=0$ a.e. in $\Omega$, a contradiction.
\end{proof}

From now on, we will make
\begin{assumption}
  \label{as:inF1}
  The target function is representable by the network, i.e., $f\in \mathcal{F}_1$.
\end{assumption}
\noindent
This means that we can take $f=f^*$ in Theorem~\ref{th:1}.

% --what
% happens when $f\not\in \mathcal{F}_1$ is briefly discussed in
% Appendix~\ref{sec:reform} in terms of the additional error
% introduced.

\subsection{Convexification at distributional level}
\label{sec:convex}

Another advantage of taken the $n\to\infty$ limit of~\eqref{eq:21} is
that it turns \eqref{eq:51} into a quadratic
objective function for~$\gamma$:
\begin{equation}
  \label{eq:21ab}
  \mathcal{L}[{\textstyle\int_D}\hat \varphi(\cdot,\zb) \gamma(d\zb)] = C_f -\int_{\hat D} \hat F(\zb) \gamma(d\zb)+\tfrac12
  \int_{\hat D\times\hat D} \hat K(\zb,\zb') \gamma(d\zb) \gamma(d\zb')
\end{equation}
This means that minimizing~\eqref{eq:21ab} over $\gamma$ rather
than~\eqref{eq:51} over $\{\thetab_i\}_{i=1}^n$ is conceptually
simpler. In particular, any minimizer $\gamma^*$ of~\eqref{eq:21ab}
solves the linear Euler-Lagrange equation:
\begin{equation}
  \label{eq:42}
  \forall \zb \in \hat D \ : \qquad \hat F(\zb) = \int_{\hat D}
  \hat K(\zb,\zb') \gamma^*(d\zb')
\end{equation}
and the loss evaluated on any $\gamma^*$ has value zero. Indeed, using
the definitions of $ \hat F$ and $\hat K$ in~\eqref{eq:22a},
\eqref{eq:42} can be written as
\begin{equation}
  \label{eq:32}
  \int_\Omega \hat \varphi(\xb,\zb) \left( f(\xb)-
    \int_{\hat D} \hat \varphi(\xb,\zb') \gamma^*(d\zb')\right)
  \nu(d\xb) =0
\end{equation}
which, by Assumptions~\ref{th:as1} and \ref{as:inF1}, has a solution such that
$f = \int_{\hat D} \hat \varphi(\cdot,\zb) \gamma^*(d\zb)$ and, as
a result, the loss evaluated on
$\int_{\hat D} \hat \varphi(\cdot,\zb) \gamma^*(d\zb)$ is zero.

\bigskip

Of course, the results above are not necessarily an assurance of convergence in practice. 
Indeed, we do not
know how to pick $\gamma\in \mathcal{M}(\hat D)$ to represent an $f\in \mathcal{F}_1$ nor can we manipulate these Radon measures explicitly:
rather we will have to learn finite $n$ approximations of the form $\gamma^{(n)} = n^{-1} \sum_{i=1}^n c_i \delta_{\zb_i}$ by
adjusting the parameters
$\{\thetab_i\}_{i=1}^n = \{c_i,\zb_i\}_{i=1}^n$ dynamically. 
Furthermore, even though the energy can be expressed
in term of $\gamma^{(n)}$, as we will see the dynamics can only be closed at the level of the empirical distribution
\begin{equation}
  \label{eq:27}
  \mu^{(n)}(dc,d\zb) = \frac1n \sum_{i=1}^n \delta
  _{c_i}(dc)\delta_{\zb_i}(d\zb) \equiv \frac1n \sum_{i=1}^n
  \delta_{\thetab_i}(d\thetab) = \mu^{(n)}(d\thetab)
\end{equation}
with $\gamma^{(n)}$ given by
\begin{equation}
  \label{eq:76}
  \gamma^{(n)} = \int_\RR c \mu^{(n)}(dc,\cdot).
\end{equation}
Viewed as a functional of $\mu\in\mathcal{M}_+(D)$ such that $\int_\RR c \mu(dc,\cdot) = \gamma \in \mathcal{M}(\hat D)$,
the loss function~\eqref{eq:21ab}
becomes
\begin{equation}
  \label{eq:64a}
  \begin{aligned}
    \mathcal{E}[\mu] &= C_f - \int_{D} F(\thetab) \mu(d\thetab) +
    \tfrac12 \int_{D\times D} K(\thetab,\thetab') \mu(d\thetab)\mu(d\thetab')\\
    & = \tfrac12 \EE_\nu\left(f - \int_{D}
      \varphi(\cdot,\thetab) \mu (d\thetab) \right)^2  \ge 0
  \end{aligned}
\end{equation}

% \section{Interacting particles with adaptive continuous charges for training}
\section{Training by gradient descent on the exact loss}
\label{sec:weighted}

Here we assume that we train the network by evolving dynamically the
parameters $\{\thetab_{\!i}(t)\}_{i=1}^n$ according to the GD
flow~\eqref{eq:34aa}, which we recall is given by the coupled ODEs,
\begin{equation}
  \label{eq:34}
      \dot\thetab_{\!i} = \nabla F(\thetab_{\!i}) -
      \frac1n\sum_{j=1}^n \nabla K(\thetab_{\!i},\thetab_{\!j}),
\end{equation}
for $i=1,\ldots, n$.  As we show in Sec.~\ref{sec:stochgrad},
\eqref{eq:34} shares many properties with the stochastic gradient
descent (SGD) used in applications, though in SGD a
multiplicative noise term persists in the equations.  The ODEs
in~\eqref{eq:34} are the GD flow on the energy:
\begin{equation}
  \label{eq:66}
  \begin{aligned}
    E(\thetab_1,\cdots,\thetab_n)= n C_f -\sum_{i=1}^n F(\thetab_i) +
    \frac1{2n} \sum_{i,j=1}^n K(\thetab_i,\thetab_j)
  \end{aligned}
\end{equation}
This energy is simply the loss function in~\eqref{eq:51} rescaled by
$n$.

\smallskip

We consider~\eqref{eq:34} with initial conditions such that every
$\thetab_{\!i}(0)$ for $i=1, \dots,n$ is drawn independently
from some probability distribution $\mu_{\text{in}}$ satisfying
\begin{assumption}
  \label{as:wellprep}
  The distribution $\mu_{\text{in}}$ is such that: (i)
  its support contains a smooth manifold that separates the regions in $D=\RR\times \hat D$ where $c>c_0$ and $c<-c_0$ for some large enough $c_0>0$;
  (ii) $\gamma_{\text{in}} = \int_{\RR} c \mu_{\text{in}}(dc,\cdot)$
  has finite total variation,
  $|\gamma_{\text{in}}|_{\text{TV}} < \infty$; and (iii) $\forall b\in \RR\ : \ \int_{\RR\times \hat D} e^{b c}\mu_{\text{in}}(dc,d\zb) <\infty.$
\end{assumption}
\noindent
Note that property (i) guarantees that $\hat \mu_{\text{in}} = \int_{\RR} \mu_{\text{in}}(dc,\cdot)$ has
  full support in $\hat D$, $\supp \hat \mu_{\text{in}} = \hat D$---we show below that this property is preserved by the dynamics. 
  Distributions $\mu_{\text{in}}$ that satisfy Assumption~\ref{as:wellprep} include e.g. 
  $$\delta_0(dc) \, \hat \mu_{\text{in}}(d\zb)\qquad \text{and} \qquad (2\pi)^{-1/2} e^{-\frac12 c^2} dc\,  \hat\mu_{\text{in}}(d\zb),$$ 
  if $\supp \hat \mu_{\text{in}}=\hat D$ in both.
We denote the measure for the infinite set
$\{\thetab_{i}(0)\}_{i\in \NN}$ constructed this way by
$\PP_{\text{in}}$. Initial conditions of this type are used
in practice.

\subsection{Empirical distribution and nonlinear Liouville equation}
\label{sec:empiricaldist}
To proceed, we consider the empirical distribution
\begin{equation}
  \label{eq:37}
  \mu_t^{(n)} = \frac1n \sum_{i=1}^n
  \delta_{\thetab_{\!i}(t)}
\end{equation}
in terms of which we can express~\eqref{eq:68aa} as
\begin{equation}
  \label{eq:68}
  f^{(n)}_t = \frac1n \sum_{i=1}^n \varphi(\cdot,\thetab_{\!i}(t))
  = \int_{D\times R} \varphi(\cdot,\thetab) \mu_t^{(n)}(d\thetab).
\end{equation}
The empirical distribution~\eqref{eq:37} is useful to work with
because it satisfies a nonlinear Liouville type equation
\begin{equation}
  \label{eq:38}
  \begin{aligned}
    \partial_t \mu_t^{(n)} & = \nabla \cdot\left( \nabla V(\thetab, [\mu_t^{(n)}])
      \mu_t^{(n)} \right)
  \end{aligned}
\end{equation}
where we defined
\begin{equation}
  \label{eq:16}
  V(\thetab, [\mu]) = -F(\thetab) + \int_{D} K(\thetab,\thetab') \mu(d\thetab')
\end{equation}
Throughout, we will interpret \eqref{eq:38} in the standard weak
sense, as in~\eqref{eq:38weak} below.  When there is a Laplacian term
in~\eqref{eq:38} this equation is called the McKean-Vlasov
equation~\cite{McKean:1966ha, dawson_large_1987, gartner_mckean-vlasov_1988,sznit:2006}; with an additional noise term added it is often
referred to as Dean's equation~\cite{Dean:1999df}.  To prove
asymptotic trainability results, we analyze the properties of the
solution to this equation as $n\to\infty$ and $t\to\infty$.

\paragraph{Derivation of~\eqref{eq:38}}
Let $\chi:D \to \RR$ be a test function, and consider
\begin{equation}
  \label{eq:85}
  \int_D \chi(\thetab)
  \mu^{(n)}_t (d\thetab)  = \frac1n \sum_{i=1}^n \chi(\thetab_i(t))
\end{equation}
Taking the time derivative of this equation and using~\eqref{eq:34}
we deduce
\begin{equation}
    \label{eq:38weak}
  \begin{aligned}
    & \int_D \chi(\thetab) \partial_t\mu_t^{(n)}(d\thetab)\\ &= \frac1n
    \sum_{i=1}^n \nabla \chi(\thetab_i(t)) \cdot \dot \thetab_i(t)\\
    & = \frac1n
    \sum_{i=1}^n \nabla \chi(\thetab_i(t)) \cdot \left (\nabla
      F(\thetab_i(t))
      -\frac1n \sum_{j=1}^n \nabla K(\thetab_i(t),\thetab_j(t))\right)\\
    & = \int_D \nabla \chi(\thetab) \cdot \left (\nabla F(\thetab)
      -\int_D \nabla K(\thetab,\thetab') \mu_t^{(n)}(d\thetab')\right)
    \mu_t^{(n)}(d\thetab)
  \end{aligned}
\end{equation}
This is the weak form of~\eqref{eq:38}.

\subsection{Law of Large Numbers (LLN)---mean field limit}
\label{sec:zero}

Since we know that, $\PP_{\text{in}}$-almost surely as $n\to\infty$,
$\mu_0^{(n)} \rightharpoonup \mu_{\text{in}}$ by the Law of Large
Numbers, we can take the limit as $n\to\infty$ of~\eqref{eq:38} to
formally deduce:

\begin{proposition}
  Let $\mu^{(n)}_t$ be given by~\eqref{eq:37} with
  $\{\thetab_i(t)\}_{i=1}^n$ the solution of~\eqref{eq:34} with initial
  condition drawn from $\PP_{\text{in}}$.  Then, as $n\to\infty$,
  $\mu_t^{(n)}\rightharpoonup \mu_t$ a.s. where $\mu_t$ satisfies
  \begin{equation}
    \label{eq:39}
    \begin{aligned}
      \partial_t \mu_t & = \nabla \cdot\left( \nabla
        V(\thetab,[\mu_t]) \mu_t \right) \qquad \mu_0 =
      \mu_{\text{in}},
    \end{aligned}
  \end{equation}
  interpreted in the weak sense.
\end{proposition}
Note that \eqref{eq:39} is the same as~\eqref{eq:38} but with
a different initial condition. Note also that~\eqref{eq:39} is the GD
flow in the Wasserstein metric~\cite{villani_optimal_2009,ambrosio_gradient_2005}. Indeed this equation can be written
as the $\tau\to0$ limit of the Jordan-Kinderlehrer-Otto (JKO) proximal scheme~\cite{Jordan:1998b}
\begin{equation}
  \label{eq:40}
  \mu_{t+\tau} \in \argmin \left(\mathcal{E}[\mu] + \tfrac12 \tau^{-1}
    W_2^2 (\mu,\mu_t)\right),\quad  \mu_0 = \mu_{\text{in}}
\end{equation}
where $W_2 (\mu,\mu_t)$ is the 2-Wasserstein distance between $\mu$
and $\mu_t$ and $\mathcal{E}[\mu]$ is defined in ~\eqref{eq:64a}.
Finally, note that the weak solutions of~\eqref{eq:39}
satisfy: for any test function $\chi:D\to \RR$,
\begin{equation}
  \label{eq:23}
  \int_D \chi(\thetab) \mu_t(d\thetab) = \int_D
  \chi(\Thetab_t(\thetab))
  \mu_{\text{in}}(d\thetab)
\end{equation}
where $\Thetab_t(\thetab)$ solves is given in terms of characteristics
\begin{equation}
  \label{eq:30}
  \dot \Thetab_t(\thetab) = -\nabla V(\Thetab_t(\thetab),[\mu_t]), \qquad
  \Thetab_0(\thetab) = \thetab.
\end{equation}
Of course, \eqref{eq:23} is not explicit since~\eqref{eq:30} depends
on $\mu_t$, but this representation formula is useful in the
sequel. In particular, notice that it implies that: (i) $\mu_t\in \mathcal{M}_+(D)$ for all $t<\infty$ since the velocity field in~\eqref{eq:30} is globally Lipschitz on $\RR\times \hat D$ by Assumption~\ref{th:ascomp} and hence the solutions to this equation exist for all $t<\infty$; and (ii) $\supp \hat \mu_t = \hat D$ with $\hat \mu_t =\int_\RR \mu_t(dc,\cdot)$ by Assumption~\ref{as:wellprep}, and $\supp \mu_t= D$ if $\supp \mu_{\text{in}}=D$.

\paragraph{The dynamics of $f_t=\lim_{n\to\infty} f^{(n)}_t$}
\label{sec:dynoff0}
We now discuss the implications of the limiting PDE for the evolution of
\begin{equation}
  \label{eq:125}
  \lim_{n\to\infty} f^{(n)}_t = \lim_{n\to\infty} \int_{D}
  \varphi(\cdot,\thetab) \mu^{(n)}_t(d\thetab)
  = \int_{D} \varphi(\cdot,\thetab) \mu_t(d\thetab) \equiv  f_t
\end{equation}
To begin, note that from \eqref{eq:16} we can express $V(\thetab,[\mu_t])$ as
\begin{equation}
  \label{eq:17b}
  V(\thetab,[\mu_t]) =
  \int_{\Omega} \left(f_t(\xb)- f(\xb) \right) \varphi(\xb,\thetab) \nu(d\xb)
\end{equation}
As a result~\eqref{eq:39} can be written as
\begin{equation}
  \label{eq:39sb}
    \partial_t \mu_t = \nabla \cdot\left( \int_{\Omega}
      \nabla_{\thetab} \varphi(\xb,\thetab) \left( f_t(\xb)
        -f(\xb)\right)\nu(d\xb)\mu_t\right)
\end{equation}
and we deduce, using~\eqref{eq:125},
\begin{equation}
  \label{eq:99lln}
  \begin{aligned}
    \partial_t f_t &= \int_{D}
    \varphi(\cdot,\thetab) \partial_t\mu_t(d\thetab)\\
    & = -\int_{D} \nabla_{\thetab}\varphi(\cdot,\thetab) \cdot\left(
      \int_{\Omega} \nabla_{\thetab} \varphi(\xb',\thetab)\left(
        f_t(\xb')
        -f(\xb')\right)\nu(d\xb')\mu_t(d\thetab)\right)\\
  \end{aligned}
\end{equation}
Interchanging the order of integration
gives:
\begin{proposition}[LLN]
  \label{th:llnft}
  Let $f^{(n)}_t$ be given by~\eqref{eq:68} with
  $\{\thetab_i(t)\}_{i=1}^n$ solution of~\eqref{eq:34} with initial
  condition drawn from $\PP_{\text{in}}$.  Then, as $n\to\infty$,
  $f_t^{(n)}\to f_t$ a.s. pointwise, where $f_t$ satisfies
\begin{equation}
  \label{eq:48}
  \partial_t f_t(\xb) = -\int_{\Omega} M([\mu_t],\xb,\xb')
  \left(f_t(\xb') - f(\xb') \right)\nu(d\xb')
\end{equation}
where we defined  the kernel
\begin{equation}
  \label{eq:96}
  \begin{aligned}
    M([\mu],\xb,\xb') &= \int_{D} \nabla_{\thetab}
    \varphi(\xb,\thetab)
    \cdot\nabla_{\thetab} \varphi(\xb',\thetab) \mu(d\thetab)\\
    &= \int_{\RR\times \hat D} \left(c^2\nabla_{\zb} \hat
      \varphi(\xb,\zb) \cdot\nabla_{\zb} \hat\varphi(\xb',\zb) +
      \hat\varphi(\xb,\zb) \hat\varphi(\xb',\zb) \right) \mu(dc,d\zb).
  \end{aligned}
\end{equation}
\end{proposition}
\noindent
The kernel~\eqref{eq:96} is symmetric in $\xb$ and $\xb'$ for any
$\mu\in \mathcal{M}(D)$ and positive semidefinite if
$\mu\in \mathcal{M}_+(D)$ since, given any $r\in L^2(\Omega,\nu)$, we
then have
\begin{equation}
  \label{eq:97}
  \begin{aligned}
    &\int_{\Omega^2} r(\xb) r(\xb') M([\mu],\xb,\xb') \nu(d\xb)
    \nu(d\xb') \\
    & = \int_{\RR\times \hat D} \left(c^2|\nabla_{\zb} R(\zb)|^2 +
      |R(\zb)|^2\right)\mu(dc,d\zb) \ge 0
  \end{aligned}
\end{equation}
where
\begin{equation}
  \label{eq:98}
  R(\zb) = \int_\Omega r(\xb) \hat\varphi(\xb,\zb) \nu(d\xb).
\end{equation}
Equation~\eqref{eq:48} also confirms that $f_t$ evolves on a
quadratic landscape, namely the loss function~\eqref{eq:21a} itself:
Indeed this equation can be written as
\begin{equation}
  \label{eq:134}
  \partial_t f_t(\xb) = -\int_{\Omega}
  M([\mu_t],\xb,\xb')\, D_{f_t(\xb')}\mathcal{L}[f_t] \nu(d\xb')
\end{equation}
where $D_{f(\xb)}$ denotes the gradient with respect to $f(\xb)$ in the
$L^2(\Omega,\nu)$-norm, i.e. given a functional $\mathcal{F}[f]$,
\begin{equation}
  \label{eq:104}
  \forall h: \Omega \to \RR \ : \
  \lim_{z\to0} \frac{d}{dz}\mathcal{F}[f+z h] = \< h, D_{f} \mathcal{F}[f]
  \>_{L^2(\Omega, \nu)} = \int_\Omega h(\xb) D_{f(\xb)} \mathcal{F}[f] \nu(d\xb).
\end{equation}
That is, $D_{f(\xb)}$ reduces to $\delta/\delta f(\xb) $ if
$\nu(d\xb) = d\xb$.

\subsection{Long time behavior---global convergence}
\label{sec:lln}

Let us now analyze the long-time solutions of~\eqref{eq:39} for the
weak limit $\mu_t$ of $\mu_t^{(n)}$ and~\eqref{eq:48} for the limit
$f_t$ of $f^{(n)}_t$.  As is well-known, \eqref{eq:39} has more
stationary points than $\mathcal{E}[\mu]$ has minimizers.  Since
\eqref{eq:39} is the Wasserstein GD flow on $\mathcal{E}[\mu]$, a
direct calculation shows that $E_t=\mathcal{E}[\mu_t]$ satisfies
\begin{equation}
  \label{eq:65}
  \frac{dE_t}{dt} = - \int_{D}
  |\nabla V(\thetab,[\mu_t])|^2\mu_t(d\thetab)
\end{equation}
This equation implies that the stationary points $\mu^s$
of~\eqref{eq:39} are the solutions of
\begin{equation}
  \label{eq:45s}
  \nabla V(\thetab,[\mu^s]) = 0 \qquad \text{for} \quad \thetab\in
  \supp \mu^s.
\end{equation}
This should be contrasted with the minimizers of
$\mathcal{E}[\mu]$, which satisfy:
\begin{equation}
  \label{eq:45}
  \left\{\begin{aligned}
    &V(\thetab,[\mu^*]) \ge \bar V[\mu^*]\qquad \text{for} \quad
    \thetab\in D\\
    &V(\thetab,[\mu^*]) = \bar V[\mu^*] \qquad \text{for} \quad
    \thetab\in \supp \mu^*
  \end{aligned}
  \right.
\end{equation}
where $\bar V[\mu] = \int_D V(\thetab,[\mu]) \mu(d\thetab)$. In
general, we cannot guarantee that the solutions to~\eqref{eq:45s} also
solve~\eqref{eq:45}. 
However, due to the specific form of the
unit, $\varphi(\xb,\thetab) = c\hat \varphi(\xb,\zb)$, the rate of decay of the energy~\eqref{eq:65} reads
\begin{equation}
  \label{eq:65b}
  \begin{aligned}
    \frac{dE}{dt} &= - \int_{\RR\times \hat D} \left(c^2 |\nabla \hat
      V(\zb,[\mu_t])|^2 + |\hat
      V(\zb,[\mu_t])|^2\right)\mu_t(dc,d\zb)\\
    & = - \int_{\RR\times \hat D} c^2 |\nabla \hat
      V(\zb,[\mu_t])|^2 \mu_t(dc,d\zb) - \int_{\hat D}  |\hat
      V(\zb,[\mu_t])|^2 \hat \mu_t(d\zb)
  \end{aligned}
\end{equation}
where $\hat \mu_t = \int_{\RR} \mu_t(dc,\cdot)$ and
\begin{equation}
  \label{eq:54}
  \hat V(\zb,[\mu]) = - \hat F(\zb) + \int_{\RR\times \hat D} c' \hat
  K(\zb,\zb') \mu(dc',d\zb')
\end{equation}
\eqref{eq:65b} implies that the stationary points $\mu^s$
of~\eqref{eq:39} satisfy
\begin{equation}
  \label{eq:24}
  \begin{aligned}
    &\hat V(\zb,[\mu^s]) = 0 \quad \text{for} \quad \zb \in \supp \hat
    \mu^s = \int_{\RR} \mu^s(dc,\cdot)\\
  \end{aligned}
\end{equation}
As a result $V(\thetab,[\mu^s]) = c\hat V(\zb,[\mu^s]) =0 $ for
$\thetab=(c,\zb) \in \supp \mu^s$, and this shows that the second
equation in~\eqref{eq:45} is automatically satisfied, noting that $\bar V
=0$ for a global minimizer.

To show that the first equation of~\eqref{eq:24} also holds, we establish that
$\hat V(\zb,[\mu^s]) = 0$ everywhere in $\hat D$. We proceed by
contradiction: Suppose that $\mu_t$ converges to some
$\mu^s$ such that $\hat V(\zb,[\mu^s]) \not= 0$ for $\zb =\hat D^c_s$
where $\hat D^c_s$ is the complement in $\hat D $ of
$\hat D_s = \supp \hat \mu^s$---the relevant case is when $\hat D^c_s$ has nonzero Hausdorff measure in $\hat D$. Looking at the characteristic
equations~\eqref{eq:30} written in terms of $\Thetab_t=(C_t,\Zb_t)$ as
\begin{equation}
  \label{eq:12}
  \left\{
    \begin{aligned}
    \dot C_t(c,\zb) &= - \hat V(\Zb_t(c,\zb),[\mu_t]), \qquad &C_0(c,\zb)=c\\
    \dot \Zb_t(c,\zb) &= -C_t(c,\zb) \nabla \hat V(\Zb_t(c,\zb),[\mu_t]),
    \qquad &\Zb_0(c,\zb)=\zb
  \end{aligned}
  \right.
\end{equation}
Since we know that
$\supp\hat \mu_t = \hat D $ at all positive time $t<\infty$, in order that
$\mu_t \to \mu^s$ as $t\to\infty$, all the mass must be expelled from $\hat D^c_s$. That is, all the solutions to~\eqref{eq:12} must leave this domain, or at
least accumulate at its boundary, while at the same time we must have
$\lim_{t\to\infty}\hat V(\zb,[\mu_t]) \not= 0$. To show that this scenario
is impossible, note that (using the fact that $\hat D$ is compact)
\begin{equation}
  \label{eq:81}
  \forall \delta >0 \quad \exists t_c>0 \ : \
  \sup_{\zb} |\hat V(\zb,[\mu_t]) - \hat V(\zb,[\mu_s])|\le \delta
  \qquad \text{if} \quad t \ge t_c.
\end{equation}
This means that, for $t\ge t_c$, to leading order in
$\delta$~\eqref{eq:12} reads
\begin{equation}
  \label{eq:12s}
  \left\{
    \begin{aligned}
    \dot C_t(c,\zb) &= - \hat
      V(\Zb_t(c,\zb),[\mu^s]), \qquad &C_0(c,\zb)=c\\
    \dot \Zb_t(c,\zb) &= -C_t(c,\zb) \nabla \hat
      V(\Zb_t(c,\zb),[\mu^s]),
    \qquad &\Zb_0(c,\zb)=\zb
  \end{aligned}
  \right.
\end{equation}
which is the GD flow on
\begin{equation}
  \label{eq:26}
  c \hat V(\zb,[\mu^s])
\end{equation}
Suppose that $\hat V(\zb,[\mu^s])>0$ somewhere in $\hat D_s^c$---the case when $\hat V(\zb,[\mu^s])<0$ somewhere in $\hat D_s^c$ can be treated similarly. Since $\hat D_s^c$ is compact, $\hat V(\zb,[\mu^s])$ must then have a maximum in $\hat D_s^c$, i.e. (using the differentiability of the unit) there exists $ \zb_1 \in\hat D_s^c$ with $\zb_1\not \in \partial \hat D_s^c$ and such that $\nabla \hat V(\zb_1,[\mu^s])=0$, $\hat V(\zb_1,[\mu^s])= \hat V_1 >0$, and  and $\hat V(\zb_1,[\mu^s])>\hat V(\zb,[\mu^s])$ for $\zb\in\hat D_s^c$. Consider the solutions to~\eqref{eq:12s} for initial $(c,\zb)$ such that 
$\Zb_t(c,\zb)$ is very close to $\zb_1$ at $t=t_c$---these 
solutions must exist since $\supp \mu_t= \hat D$ for all $t<\infty$. If among these solutions there are some such that $C_t(c,\zb)$ is negative at time $t=t_c$ (which is always the case if $\supp \mu_{\rm in}=D$ since $\supp \mu_t = D$ for all $t<\infty$ in that case), then by~\eqref{eq:12s} $C_t(c,\zb)$ becomes more negative  and $\Zb_t(c,\zb)$ gets closer to $\zb_1$ for $t>t_c$. If all $C_{t+\tau}(c,\zb)$ are positive at time $t=t_c(\delta)$, then the corresponding $\Zb_t(c,\zb)$ go away from $\zb_1$ for as long as their $C_t(c,\zb)$ remains positive; however, eventually some $C_t(c,\zb)$ become negative (since $C_{t+\tau}(c,\zb_1) = C_{t}(c,\zb_1)- \tau\hat V(\zb_1,[\mu^s])= C_{t}(c,\zb_1)- \tau\hat V_1$ under~\eqref{eq:12s}), at which point we go back to the first case and $\Zb_t(c,\zb)$ gets closer to $\zb_1$. Either way, we can always find solutions with $\Zb_t(c,\zb)$ sufficiently close to $\zb_1$ at time $t_c(\delta)$ that will eventually converge to $\zb_1$ rather than exiting $\hat D_s^c$, a contradiction with our assumption that all solutions must either exit this domain or accumulate at its boundary. This argument is based on~\eqref{eq:12s} rather than the original~\eqref{eq:12}, but by setting $\delta$ small enough (and $t_c$ large enough) we can make the terms left over in~\eqref{eq:12s} arbitrarily small so that they do not affect the result.

This concludes the justification that the stationary points $\mu^s$ of~\eqref{eq:39} are such that $\hat V(\zb,[\mu^s])=0$ everywhere in $\hat D$, i.e. they are minimizers of $\mathcal{E}[\mu]$, which from~\eqref{eq:24} implies
\begin{equation}
  \label{eq:44}
  \forall \zb \in \hat D \ : \qquad 0= \int_{\Omega} \hat \varphi(\xb,\zb)\left( f(\xb) - \int_{\hat
      D}\hat\varphi(\xb,\zb')\gamma^s(d\zb') \right) \nu(d\xb)
\end{equation}
where $\gamma^s = \int_{\RR} c \mu^s(dc,\cdot)$. As a result, by
Assumptions~\ref{th:as1} and~\ref{as:inF1},
\begin{equation}
  f =\int_{\hat D}\hat\varphi(\cdot,\zb)\gamma^s(d\zb),
  \label{eq:56}
\end{equation}
In other other words, we have established:
\begin{proposition}[Global convergence]
  \label{thm:limrhot}
  Let $\mu_t$ be the solution to~\eqref{eq:39} for the initial
  condition $\mu_0=\mu_{\text{in}}$ that satisfy
  Assumption~\ref{as:wellprep}.  If
  $\mu_t \to \mu^*\in \mathcal{M}_+(D)$ as $t\to\infty$, then under
  Assumptions~\ref{th:as1} and~\ref{as:inF1} $\mu^*$
  is a minimizer of $\mathcal{E}[\mu]$ and we have
  \begin{equation}
    \label{eq:31}
    \lim_{t\to\infty} \int_{D} \varphi(\cdot, \thetab)
    \mu_t(d\thetab) =\int_{D} \varphi(\cdot, \thetab)
    \mu^*(d\thetab)= f.
  \end{equation}
\end{proposition}
\noindent
Note that we assume that $\mu_t$ converges to some probability
measure to establish this proposition. This is because we cannot
exclude \textit{a~priori} that the dynamics eventually loses mass at
infinity, e.g. if some of the solutions of the characteristic
equation~\eqref{eq:30} eventually diverge as $t\to\infty$. We do not expect this
scenario to occur for most initial conditions and one way to preclude it
altogether is to add regularizing terms in the loss function.

The argument that leads to~\eqref{eq:31} would be simple if it were the case that
$\hat \mu^*= \int_{\RR} \mu^*(dc,\cdot)$ has full support in $\hat D$. 
Indeed this would imply that the kernel~\eqref{eq:96} evaluated on $\mu_t$ is positive
definite not only for all $t\ge 0$ but also in the limit as
$t\to\infty$ and hence the only fixed point of~\eqref{eq:48} is $f$. 
It is reasonable to assume that $\supp \hat \mu^*=\hat D$ because: (i)
$\supp \hat \mu_t = \hat D$ for all $t<\infty$ as mentioned before
and (ii) there is no energetic incentive to shrink the support, even when
$t\to\infty$. This see why, note that if $\mu^*$ is an energy
minimizer such that $\supp \hat \mu^*\not =\hat D$, then a direct
calculation shows that for any $\alpha\in(0,1)$ and any
$\hat \mu\in \mathcal{M}_+(\hat D)$ with $\supp \hat \mu = \hat D$,
\begin{equation}
  \label{eq:63}
  \mu^{**}(dc,d\zb) = (1-\alpha)^2 \mu^*((1-\alpha) dc, d\zb) + \alpha
  \delta_0(dc) \hat \mu(d\zb)
\end{equation}
is also a energy minimizer in $\mathcal{M}_+(D)$ such that
$\hat \mu^{**} = \int_{\RR} \mu^{**}(dc,\cdot)$ has support $\hat
D$.

In Appendix, we analyze the behavior of
$\mu_t$ on a longer timescale and show that, with noise and certain
regularizing terms added in~\eqref{eq:34}, $\mu_t$ reaches a unique
fixed point $\mu^*\in \mathcal{M}_+(D)$ such that
$\int_{D} \log (d\mu^*/d\mu^0) d\mu^* <\infty$, where $\mu^0$ is some
prior measure used for regularization.

\bigskip

We can summarize the results of Secs.~\ref{sec:zero} and \ref{sec:lln}
into:

\begin{proposition}[LLN \& global convergence]
  \label{th:lln}
  Let $f^{(n)}_t$ be given by~\eqref{eq:68} with
  $\{\thetab_i(t)\}_{i=1}^n$ solution of~\eqref{eq:34} with
  initial condition drawn from $\PP_{\text{in}}$. Then under the
  conditions of Proposition~\ref{thm:limrhot} we have
  \begin{equation}
    \label{eq:86lln}
    \lim_{n\to\infty} f^{(n)}_t = f_t
    \qquad \text{pointwise, \  $\PP_{\text{in}}$-almost surely}
  \end{equation}
where $f_t$ solves~\eqref{eq:48} and satisfies
\begin{equation}
  \label{eq:127}
  \lim_{t\to\infty} f_t = f \quad \text{a.e.\ \ in  \ \ $\Omega$}.
\end{equation}
\end{proposition}
\noindent
The convergence in~\eqref{eq:127} is equivalent to the statement in
Proposition~\ref{thm:limrhot}.  Notice that, since the evolution of
$f_t$ occurs via~\eqref{eq:48}, which is independent of~$n$,  for any $\delta>0$ we can find $t_c$ independent of $n$ such that for $t>t_c$, $\EE_\nu|f_t-f|^2<\delta$. Since for any $\delta >0$ and $t>0$ we can also find $n_c$ such that for $n>n_c$,  $\EE_\nu|f^{(n)}_t-f_t|^2<\delta$, we can
interchange the limits in $n$ and $t$ in Theorem~\ref{th:lln}, i.e. we
also have
\begin{equation}
  \label{eq:15}
  \lim_{n\to\infty} \lim_{t\to\infty} f^{(n)}_t = f.
\end{equation}

\subsection{Central Limit Theorem (CLT)}
\label{sec:clt}

Let us now consider the fluctuations of $\mu_t^{(n)}$ around its limit
$\mu_t$. To this end, we define~$\omega^{(n)}_t$ via:
\begin{equation}
  \label{eq:74}
  \omega^{(n)}_t = n^{1/2} \left( \mu_t^{(n)}-\mu_t\right),
\end{equation}
Explicitly, \eqref{eq:74} means:
\begin{equation}
  \label{eq:106}
  \omega^{(n)}_t  = n^{-1/2} \sum_{i=1}^n
  \left(\delta_{\thetab_{i}(t)} - \mu_t\right)
\end{equation}
The scaling factor $n^{1/2}$ is set by the fluctuations in the initial
conditions: if we pick a test function $\chi:D\to\RR$ the
CLT tells us that under $\PP_{\text{in}}$
\begin{equation}
  \label{eq:112init}
  \begin{aligned}
    \int_{D} \chi(\thetab) \omega^{(n)}_0(d\thetab)  =
    n^{-1/2} \sum_{i=1}^n \tilde \chi(\thetab_{i}(0)) \to
    N(0,C_\chi) \quad \text{in law as\ \  $n\to\infty$}
  \end{aligned}
\end{equation}
where
$\tilde \chi(\thetab) = \chi(\thetab) -\int_{D} \chi(\thetab)
\mu_{\text{in}}(d\thetab)$ and $N(0,C_\chi) $ denotes the Gaussian
random variable with mean zero and variance
\begin{equation}
  \label{eq:135}
  C_\chi = \int_{D} \left |\tilde \chi(\thetab)\right|^2
  \mu_{\text{in}}(d\thetab),
\end{equation}
To see what happens at later times, we derive an equation
for~$\omega^{(n)}_t$ by subtracting~\eqref{eq:39}
from~\eqref{eq:38} and using~\eqref{eq:74}
\begin{equation}
  \label{eq:38om}
  \partial_t \omega^{(n)}_t
  = \nabla \cdot\left( \omega^{(n)}_t \nabla
    V(\thetab,[\mu_t]) + \left(\mu_t + n^{-1/2} \omega_t^{(n)} \right)\nabla F(\thetab,[\omega^{(n)}_t])\right)
\end{equation}
where we defined
\begin{equation}
  \label{eq:46}
  F(\thetab,[\mu])  = \int_{D} K(\thetab,\thetab') \mu(d\thetab')
\end{equation}
If we take the limit as $n\to\infty$, the term involving
$ n^{-1/2} \omega_t^{(n)} $ at the right hand side
of~\eqref{eq:38om} disappears (we quantify its rate of
convergence to zero in more detail in Sec.~\ref{sec:first}) and we
formally deduce that
\begin{proposition}
  Let $\omega^{(n)}_t$ be given by~\eqref{eq:106} with
  $\{\thetab_i(t)\}_{i=1}^n$ solution of~\eqref{eq:34} with initial
  conditions draw from $\PP_{\text{in}}$ and $\mu_t$ solution
  to~\eqref{eq:39sb}. Then
  \begin{equation}
    \label{eq:33}
    \omega_t^{(n)} \rightharpoonup
    \omega _t \quad \text{in law as \ \  $n\to \infty$}
  \end{equation}
  where $\omega _t$ satisfies
  \begin{equation}
    \label{eq:38omlim}
    \begin{aligned}
      \partial_t \omega_t = \nabla \cdot\left( \omega_t \nabla
        V(\thetab,[\mu_t]) + \mu_t \nabla F(\thetab,[\omega_t])\right)
    \end{aligned}
  \end{equation}
  to be solved in the weak sense with the Gaussian initial conditions
  read from~\eqref{eq:112init}:
  \begin{equation}
    \label{eq:1}
    \int_{D} \chi(\thetab) \omega_0(d\thetab) =
    N(0,C_\chi)
  \end{equation}
\end{proposition}
Note that since the mean of $\omega_0$ is zero initially
and~\eqref{eq:38omlim} is linear, this mean remains zero for all
times, and we can focus on the evolution of its covariance:
\begin{equation}
  \label{eq:143}
  \Sigma_t(d\thetab,d\thetab') = \EE_{\text{in}} [ \omega_t(d\thetab)
  \omega_t(d\thetab') ]
\end{equation}
From~\eqref{eq:38omlim} it satisfies
\begin{equation}
  \label{eq:38tildeG1}
  \begin{aligned}
    \partial_t \Sigma_t& = \nabla \cdot\left( \Sigma_t \nabla
      V(\thetab,[\mu_t]) + \mu_t(d\thetab)
      \nabla G(\thetab,d\thetab',[\Sigma_t])\right)\\
    & + \nabla' \cdot\left( \Sigma_t \nabla
      V(\thetab',[\mu_t]) + \mu_t(d\thetab')
      \nabla G(\thetab', d\thetab,[\Sigma_t])\right)
  \end{aligned}
\end{equation}
where we defined
\begin{equation}
  \label{eq:64}
  G(\thetab,\cdot,[\Sigma])) = \int_{D}  K(\thetab,\thetab'') \Sigma(d\thetab'',\cdot)
\end{equation}
Equation~\eqref{eq:38tildeG1} should be interpreted in the weak sense
and solved for the initial condition
\begin{equation}
  \label{eq:144}
  \Sigma_0(d\thetab,d\thetab')=
  \mu_{\text{in}}(d\thetab) \delta_{\thetab}(d\thetab')
\end{equation}

\medskip

\paragraph{The dynamics of
  $g_t = \lim_{n\to\infty} n^{1/2} ( f^{(n)}_t - f_t)$} We can also
test these equations against the unit, to deduce that, as
$n\to\infty$,
\begin{equation}
  \label{eq:73}
  \begin{aligned}
    g^{(n)}_t & = \int_D
    \varphi(\cdot, \thetab) \omega^{(n)}_t(d\thetab) = n^{1/2} \big(f^{(n)}_t - f_t\big)  \\
    & = n^{-1/2} \sum_{i=1}^n \left(\varphi(\cdot,\thetab_{i}(t))-
      f_t\right)
  \end{aligned}
\end{equation}
converges in law, $g^{(n)}_t\to g_t$, where $g_t$ is a Gaussian
process satisfying
\begin{equation}
  \label{eq:75}
  \begin{aligned}
    \partial_t g_t &= -\int_\Omega M(\xb,\xb',[\omega_t])
    \left(f_t(\xb')-f(\xb')\right)\nu(d\xb') \\
    & \quad - \int_\Omega M(\xb,\xb',[\mu_t]) g_t(\xb')\nu(d\xb')
  \end{aligned}
\end{equation}
This equation should be solved with Gaussian initial conditions with
mean zero and covariance
\begin{equation}
  \label{eq:49}
  \begin{aligned}
    C_0(\xb,\xb') = \EE_{\text{in}} [g_0(\xb) g_0(\xb') ] &= \int_{D}
    \varphi(\xb,\thetab) \varphi(\xb',\thetab)
    \mu_{\text{in}}(d\thetab)\\
    &- \int_{D\times D} \varphi(\xb,\thetab) \varphi(\xb',\thetab')
    \mu_{\text{in}}(d\thetab) \mu_{\text{in}}(d\thetab')
  \end{aligned}
\end{equation}
Since~\eqref{eq:75} is linear the mean of $g_t$ remains zero at all
times and we can again focus on the evolution of its covariance:
\begin{equation}
  \label{eq:53}
  C_t(\xb,\xb') =\EE_{\text{in}} [g_t(\xb) g_t(\xb') ]
\end{equation}
We obtain
\begin{equation}
  \label{eq:86}
  \begin{aligned}
    \partial _t C_t & = -\int_\Omega N(\xb,\xb',\xb'', [\Sigma_t])
    \left(f_t(\xb'')-f(\xb'')\right)\nu(d\xb'')\\
    & \quad - \int_\Omega M(\xb,\xb'',[\mu_t])
    C_t(\xb'',\xb)\nu(d\xb'') \\
    &\quad - \int_\Omega M(\xb',\xb'',[\mu_t])
  C_t(\xb'',\xb')\nu(d\xb'')
  \end{aligned}
\end{equation}
where $\Sigma_t$ solves~\eqref{eq:38tildeG1} and
\begin{equation}
  \label{eq:88}
  \begin{aligned}
    N(\xb,\xb'\xb'', [\Sigma]) & = \int_{D\times D}
      \nabla_{\thetab} \varphi(\xb,\thetab) \cdot \nabla_{\thetab}
      \varphi(\xb'',\thetab) \varphi(\xb',\thetab')  \Sigma(d\thetab,d\thetab')\\
      & + \int_{D\times D} \nabla_{\thetab}
      \varphi(\xb',\thetab) \cdot \nabla_{\thetab}
      \varphi(\xb'',\thetab) \varphi(\xb,\thetab')
    \Sigma(d\thetab,d\thetab')
  \end{aligned}
\end{equation}
Summarizing, we have established:

\begin{proposition}[CLT]
  \label{th:cltg}
  Let $g^{(n)}_t$ be given by~\eqref{eq:73} with
  $\{\thetab_i(t)\}_{i=1}^n$ solution of~\eqref{eq:34} with initial
  conditions draw from $\PP_{\text{in}}$ and $\mu_t$ solution
  to~\eqref{eq:39sb}. Then, as $n\to\infty$, $g^{(n)}_t\to g_t$ in
  law, where $g_t$ is the zero mean Gaussian process whose covariance
  solves to~\eqref{eq:86} for the initial condition~\eqref{eq:49}.
\end{proposition}
\noindent
%We will bound the covariance of $g_t$ next.

\subsection{Scaling of the fluctuations at long and very long times}
\label{sec:first}

To analyze the behavior of the fluctuations as $t\to\infty$,
we revisit the results from the last section from a
different perspective. Suppose that, instead of~\eqref{eq:106}
and~\eqref{eq:73}, we would consider
\begin{equation}
  \label{eq:79}
  \bar \omega^{(n)}_t = n^{-1/2} \sum_{i=1}^n \left(\delta_{\Thetab_{i}(t)}-
   \mu_t\right)
\end{equation}
and
\begin{equation}
  \label{eq:79g}
  \bar g^{(n)}_t  = n^{-1/2} \sum_{i=1}^n \left(\varphi(\cdot,\Thetab_{i}(t))-
    \int_D \varphi(\cdot,\theta) \mu_t(d\thetab)\right)
\end{equation}
where $\Thetab_{i}(t)$ are independent copies of the mean-field
characteristic equation~\eqref{eq:30}. Then, direct calculations show
that $\bar \omega^{(n)}_t\rightharpoonup \bar\omega_t$ and
$\bar g^{(n)}_t\to\bar g_t$ in law as $n\to\infty$, where $\bar\omega_t$ and
$\bar g_t$  are Gaussian processes with mean zero and covariance given
explicitly by
\begin{equation}
  \label{eq:80}
  \bar \Sigma_t(d\thetab,d\thetab') =
  \EE_{\text{in}} [ \bar\omega_t (d\thetab) \bar\omega_t(d\thetab')] =
  \mu_t(d\thetab) \delta_{\thetab} (d\thetab') -
  \mu_t(d\thetab) \mu_t(d\thetab')
\end{equation}
and
  \begin{equation}
  \label{eq:80C}
  \bar C_t(\xb,\xb') =
  \EE_{\text{in}} [ \bar g_t (\xb) \bar g_t(\xb')] =
  \int_D \varphi(\xb,\thetab)\varphi(\xb',\thetab) \mu_t(d\thetab) -
  f_t(\xb) f_t(\xb')
\end{equation}
We can also easily write down evolution equations for $\bar\omega_t$
and $\bar g_t$: they read
 \begin{equation}
   \label{eq:38omlimmf}
   \partial_t \bar\omega_t = \nabla \cdot\left( \bar\omega_t \nabla
     V(\thetab,[\mu_t]) \right)
 \end{equation}
 and
 \begin{equation}
  \label{eq:75mf}
    \partial_t \bar g_t = -\int_\Omega M(\xb,\xb',[\bar \omega_t])
    \left(f_t(\xb')-f(\xb')\right)\nu(d\xb')
\end{equation}
Let us focus on this last equation: it is similar to \eqref{eq:75}, but without the last term,
$-\int_\Omega M(\xb,\xb',[\mu_t]) g_t(\xb') \nu(d\xb')$. 
Since the kernel $M$ is positive semi-definite, we know that
the solutions to \eqref{eq:75} are controlled by those of
\eqref{eq:75mf}. In particular, 
\begin{equation}
  \label{eq:91}
  \EE_{\text{in}} \int_{\Omega} |g_t(\xb)|^2 \nu(d\xb) = \int_{\Omega}
  C_t(\xb,\xb) \nu(d\xb) \le \int_{\Omega}
  \bar C_t(\xb,\xb) \nu(d\xb)
\end{equation}
If we assume that $\mu_t\to\mu^*\in \mathcal{M}_+(D)$ as
$t\to\infty$, from~\eqref{eq:80C} we have
\begin{equation}
  \label{eq:93}
  \lim_{t\to\infty} \int_{\Omega}
  \bar C_t(\xb,\xb) \nu(d\xb) = \int_{D} K(\thetab,\thetab)
  \mu^*(d\thetab) - \int_\Omega |f(\xb)|^2 \nu(d\xb)
\end{equation}
and therefore
\begin{equation}
  \label{eq:91lim}
  \lim_{t\to\infty}  \int_{\Omega} C_t(\xb,\xb)
  \nu(d\xb) \le \int_{D} K(\thetab,\thetab)
  \mu^*(d\thetab) - \int_\Omega |f(\xb)|^2 \nu(\d\xb).
\end{equation}
Because
%q
\begin{equation}
  \label{eq:100}
   \int_{\Omega} C_t(\xb,\xb) \nu(d\xb)= \lim_{n\to\infty}
  n \EE_{\text{in}}\int_{\Omega} |f^{(n)}_t(\xb)-f_t(\xb)|^2\nu(d\xb)
\end{equation}
the previous result gives a Monte-Carlo type error bound on the loss. Note that this bound is only nontrivial if
\begin{equation}
\label{eq:condonbound}
\begin{aligned}
\int_{D} K(\thetab,\thetab) \mu^*(d\thetab) 
& = \int_{\RR \times \hat D } c^2 \hat K(\zb,\zb) \mu^*(dc,d\zb) \\
& \le \|\hat K\|_\infty \int_{\RR \times \hat D } c^2 \mu^*(dc,d\zb) < \infty
\end{aligned}
\end{equation}
During training, we have $\int_{\RR \times \hat D } c^2 \mu_t(dc,d\zb)<\infty$ for all $t<\infty$, and to guarantee that this moment does not blow up as $t\to\infty$, or more generally to control its value in that limit, we may need to add a regularizing term to the loss function. If~\eqref{eq:condonbound} holds,
there is a situation in which we can even deduce a better bound: if
$\supp \hat \mu^* = \hat D$ and , then $M(\xb,\xb,[\mu_t])$ is positive
definite for all $t\ge0$ and in the limit $t\to\infty$, indicating
that the last term in~\eqref{eq:75} is always dissipative. In this
case the argument above shows that
\begin{equation}
  \label{eq:103}
  \lim_{t\to\infty}  \int_{\Omega} C_t(\xb,\xb)
  \nu(d\xb)  = 0.
\end{equation}
Summarizing:

\begin{proposition}[Fluctuations at long times]
  \label{th:cltglt}
  Let $f^{(n)}_t$ be given by with $\{\thetab_i(t)\}_{i=1}^n$ solution
  of~\eqref{eq:68} with initial conditions draw from $\PP_{\text{in}}$
  and $f_t$ solution to~\eqref{eq:48}. Then, under the conditions
  of Proposition~\ref{thm:limrhot} and assuming that~\eqref{eq:condonbound} holds, we have
  \begin{equation}
    \label{eq:105}
    \begin{aligned}
      & \lim_{t\to\infty} \lim_{n\to\infty} n\, \EE_{\text{in}}
      \int_{\Omega} |f^{(n)}_t(\xb)-f_t(\xb)|^2\nu(d\xb)\\
      & \quad \le \int_{D} K(\thetab,\thetab) \mu^*(d\thetab) -
      \int_\Omega |f(\xb)|^2 \nu(d\xb)
    \end{aligned}
  \end{equation}
  In addition, if $\supp\hat \mu^* = \hat D$, we have
  \begin{equation}
    \label{eq:105b}
    \begin{aligned}
      & \lim_{t\to\infty} \lim_{n\to\infty} n \, \EE_{\text{in}}
      \int_{\Omega} |f^{(n)}_t(\xb)-f_t(\xb)|^2\nu(d\xb)=0.
    \end{aligned}
  \end{equation}
\end{proposition}

\bigskip

In situations where $\supp\hat \mu^* = \hat D$ and~\eqref{eq:105b} holds, we
see that the fluctuations, initially detectable on the scale $n^{-1/2}$, become higher order as
time increases. 
To understand the scale at which the fluctuations eventually settle, consider
\begin{equation}
  \label{eq:106lt}
  \tilde \omega^{(n)}_t (d\thetab) = n^{\xi(t)} \sum_{i=1}^n
  \left(\delta_{\thetab_{i}(t)}(\thetab) - \mu_t(d\thetab)\right)
\end{equation}
where $\xi(t)$ is some time-dependent exponent to be specified. By
proceeding as we did to derive~\eqref{eq:38om}, we have that
$\tilde \omega^{(n)}_t$ satisfies
\begin{equation}
  \label{eq:38tildelt}
  \begin{aligned}
    \partial_t \tilde \omega^{(n)}_t & = \nabla \cdot\left(
      \tilde\omega^{(n)}_t \nabla V(\thetab,[\mu_t]) + \mu_t \nabla
      F(\thetab,[\tilde\omega^{(n)}_t])\right)\\
    & + n^{-\xi(t)} \nabla \cdot\left(\omega_t^{(n)} \nabla
      F(\thetab,[\tilde\omega^{(n)}_t])\right) + \dot \xi(t) \log n \,
    \tilde \omega^{(n)}_t.
  \end{aligned}
\end{equation}
In order to take the limit as $n\to\infty$ of this equation, we need
to consider carefully the behavior of the factors
in~\eqref{eq:38tildelt} that contain $n$ explicitly, that is,
$\omega_t^{(n)} \nabla F(\thetab,[\tilde\omega^{(n)}_t])$ and
$\dot \xi(t) \log n \, \tilde\omega^{(n)}_t$. Regarding the former, 
for any $p\in \NN$ and $\xi\in \RR$,
\begin{equation}
  \label{eq:87}
  \begin{aligned}
    & \EE_{\text{in}} \left(n^{-\xi} \int_{D\times D}
      \chi(\thetab)\chi(\thetab') \tilde\omega^{(n)}_0(d\thetab)
    \tilde\omega^{(n)}_0(d\thetab')\right)^p = O\left(n^{(\xi-1)p}\right),
  \end{aligned}
\end{equation}
which can be verified by a direct calculation. 
For example if $p=1$, this expectation is $n^{(\xi-1)} C_\chi$ where
$C_\chi$ is given in~\eqref{eq:135}. 
Equation~\eqref{eq:87} implies
that
$n^{-\xi} \tilde\omega^{(n)}_0(d\thetab)
\tilde\omega^{(n)}_0(d\thetab)\rightharpoonup 0$ weakly in $L^{2p}$ at
$t=0$ for any $\xi<1$. To see whether we can bring the fluctuations to
that scale, notice that if we set
\begin{equation}
  \label{eq:3}
  \dot \xi(t) \log n  = o(1)
\end{equation}
the last term at the right hand side of~\eqref{eq:38tildelt} is also
higher order---\eqref{eq:3} means that we can vary $\xi(t)$, but only
slowly. \eqref{eq:3} can be achieved by choosing e.g.
\begin{equation}
\label{eq:admixi}
\xi(t) =     \bar \xi(t/a_n) 
\end{equation}
with $\bar \xi(0) =
    \tfrac12$, $\bar\xi'(u) >0$, $\lim_{u\to\infty} \bar\xi(u)
    =  <1$, and $a_n$ growing with 
    $n$ and such that $\lim_{n\to\infty} a_n/\log n =\infty$.
With this choice, both the last two terms at the right hand
of~\eqref{eq:38tildelt} are a small perturbation that vanishes as
$n\to\infty$. Therefore, if we test $\tilde \omega^{(n)}_t$ against the unit, and
define
\begin{equation}
  \label{eq:73lt}
  \tilde g^{(n)}_t = n^{\xi(t)} \big(f^{(n)}_t - f_t\big) = \int_D \varphi(\cdot,
  \thetab) \tilde \omega^{(n)}_t(d\thetab)
\end{equation}
we know that, if~$\xi(t)$ is as in~\eqref{eq:admixi} and $\supp\hat\mu^* = \hat D$, this
field will be controlled and go to zero eventually.  Summarizing we
have established:

\begin{proposition}[Fluctuations at very long times]
  \label{th:cltgvlt} Assume that the conditions of
  Proposition~\ref{th:cltglt} hold and $\supp\hat\mu^* = \hat D$. Then
  \begin{equation}
    \label{eq:86clt}
    \forall \xi < 1 \quad : \quad  \lim_{n\to\infty} n^{2\xi} \EE_{\text{in}}
      \int_{\Omega} |f^{(n)}_{a_n}(\xb)-f(\xb)|^2\nu(d\xb)=0
  \end{equation}
  if $a_n$ grows with $n$ and is such that
  $\lim_{n\to\infty} a_n/\log n =\infty$.
\end{proposition}

\noindent This proposition can be stated as~\eqref{eq:55zT}. It shows
a remarkable self-healing property of the dynamics: the fluctuations
at scale $O(n^{-1/2})$ of $f^{(n)}_t$ around $f_t$ that were present
initially decrease in amplitude as time progresses, and become
$O(n^{-1})$ or smaller as $t\to\infty$.

\section{Training by online stochastic gradient descent}
\label{sec:stochgrad}

In most applications, it is not possible to evaluate the expectation over the data
in~\eqref{eq:22a} defining $\hat F(\zb)$ and $\hat K(\zb,\zb')$. This is
especially true for $\hat F(\zb)$, since we typically have limited access
to $f(\xb)$: often, we can only evaluate it pointwise or only know its value on a discrete set of
points. In these cases,  we typically need to
approximate the expectation in~\eqref{eq:22a} by a sum over a finite subset
of $\xb$'s obtained by sampling from the measure $\nu$.  

If we were to fix this training data set, $\{\xb_p\}_{p=1}^P$, and denote by $\nu_P = P^{-1}\sum_{p=1}^P \delta_{\xb_p}$ the corresponding empirical measure, then all the results in Sec.~\ref{sec:weighted} apply at empirical level if we replace everywhere $\nu$ by $\nu_P$. This, however, is not the question we are typically interested in, which is rather: 

\begin{quote}
    {\it How does the test error  (that is, the error obtained using the exact loss defined with the original $\nu$) scale if we train the network on the empirical loss associated to $\nu_P$?}
\end{quote}

Here we will address this question in the specific setting of ``online'' learning algorithms, in which we can draw a training data set of batch size $P$ at every step of the learning. 
This effectively assumes that we have access to infinite data, but  cannot use it all at the same time, and the finite size of the batch 
introduces noise into the learning algorithm.
%If this learning is done using GD on the empirical $\hat F_P$ and $\hat K_P$ instead of the exact ones, the resulting scheme is referred to as
%stochastic gradient descent (SGD).  --No, GD on F_P/K_P is GD. SGD is mini-batch sampling
The algorithm in which the gradient is estimated from a subset of training data at each step is known as stochastic gradient descent.
It reads
\begin{equation}
  \label{eq:34discrete2}
      \hat \thetab_{i}(t+\Delta t) = \hat \thetab_{i}(t) + \nabla
      F_{P}(t,\hat\thetab^P_{i}(t) ) \Delta t - \frac1n\sum_{j=1}^n
      \nabla K_P(t,\hat\thetab_{i}(t),\hat\thetab_{j}(t)) \Delta t
  %\right.
\end{equation}
where $i=1,\ldots,n$, $\Delta t>0$ is some time-step, and we defined
\begin{equation}
  \begin{aligned}
  \label{eq:47}
  F_{P} (t,\thetab) &= \frac1P \sum_{p=1}^P
  f(\xb_{p}(t))\varphi(\xb_{p}(t),\thetab), \\
  K_P (t,\thetab,\thetab') &= \frac1P \sum_{p=1}^P
  \varphi(\xb_{p}(t),\thetab) \varphi(\xb_{p}(t),\thetab')
\end{aligned}
\end{equation}
in which $\{\xb_{p}(t)\}_{p=1}^P$ are $P$ iid variables which are
redrawn from $\nu$ independently at every time step~$t$. Next we analyze how the result from Sec.~\ref{sec:weighted} must be modified when we use~\eqref{eq:47} rather than \eqref{eq:34} to perform the training.

\subsection{Limiting stochastic differential equation}
\label{lim:sde}

To analyze the properties of~\eqref{eq:34discrete2}, we start start by noticing that the term
\begin{equation}
      \Rb_i(\vec\thetab) = \nabla F_{P}(t,\thetab_{i} ) - \frac1n\sum_{j=1}^n
      \nabla K_P(t,\thetab_{i},\thetab_{j}), \qquad \vec\thetab=(\thetab_1, \ldots,\thetab_n) 
\end{equation}
is an unbiased estimator of the right hand side of the GD equation~\eqref{eq:34}. Indeed, conditional on $\{\thetab_i\}_{i=1}^n$ fixed, we have
\begin{equation}
      %\EE_\nu \Rb_i(\vec\thetab) = F(t,\thetab_{i} ) - \frac1n\sum_{j=1}^n
      %\nabla K(t,\thetab_{i},\thetab_{j})
      \EE_\nu \Rb_i(\vec\thetab) = \nabla F(\thetab_{i} ) - \frac1n\sum_{j=1}^n
      \nabla K(\thetab_{i},\thetab_{j})
\end{equation}
This means that, if we split the right hand side of~\eqref{eq:34discrete2} into its expectation plus a zero-mean fluctuation, the expression resembles
an Euler-Maruyama scheme for a stochastic differential equation (SDE), except that the scaling of the
noise term involves $\Delta t$ rather than
$\sqrt{\Delta t}$. To write this SDE explicitly, we compute the covariance of $\Rb(\vec\thetab)$ conditional on $\{\thetab_i\}_{i=1}^n$ fixed, 
\begin{equation}
    \cov_\nu[ \Rb_i(\vec\thetab),\Rb_j(\vec\thetab')] = A([f-f^{(n)}],\thetab_i,\thetab_j')
\end{equation}
where $f^{(n)}= n^{-1} \sum_{i=1}^n \varphi(\cdot,\thetab_i)$ and we defined
\begin{equation}
  \begin{aligned}
  \label{eq:161}
    A([f],\thetab,\thetab') & = \EE_\nu [|f|^2
    \nabla_{\thetab}\varphi(\cdot,\thetab) \otimes \nabla_{\thetab'}\varphi(\cdot,\thetab')]\\
    & - \EE_\nu [f \,\nabla_{\thetab}\varphi(\cdot,\thetab)] \otimes  \EE_\nu 
    [f \, \nabla_{\thetab'}\varphi(\cdot,\thetab')]
  \end{aligned}
\end{equation}
The SDE capturing the behavior of the solution to~\eqref{eq:34discrete2} is
\begin{equation}
  \label{eq:34trainz}
      d\thetab_{i} = \nabla F(\thetab_{i}) dt -
      \frac1n\sum_{j=1}^n \nabla K(\thetab_{i},\thetab_{j})
      dt +
      \sqrt{\sigma}\, d\Bb_i,
\end{equation}
where $\sigma = \Delta t/P$ and $\{d\Bb_i\}_{i=1}^n$ is a white-noise process with
quadratic variation
\begin{equation}
  \label{eq:59}
    \< d\Bb_i,d\Bb_j\> =
    A([f-f^{(n)}],\thetab_{i},\thetab_{j}) dt.
\end{equation}
More precisely~\cite{Li:2015,Hu:2017uj},
\begin{lemma}
  \label{th:lem1}
  Given given any test functions $\chi:D\to\RR$ and any $T>0$, there is a constant $C>0$ such that
  \begin{equation}
    \label{eq:92}
      \sup_{0\le k\Delta t\le T} \Big|\frac1n \sum_{i=1}^n \left(\EE  \chi(\hat\theta_i(k\Delta t)) - \EE  \chi(\thetab_i(k\Delta t))
        \right) \Big| \le C \Delta t.
  \end{equation}
  where $\hat \thetab_i(t)$ and $\thetab_i(t)$ denote the solutions to~\eqref{eq:34discrete2} and \eqref{eq:38trainz}, respectively.
\end{lemma}
\noindent
This lemma is a direct consequence of the fact
that~\eqref{eq:34discrete2} can be viewed as the Euler-Maruyama
discretization scheme for~\eqref{eq:34trainz}, and this scheme has weak
order of accuracy $1$.  Note that if we let $\Delta t \to0$,
\eqref{eq:34trainz} reduces to the ODEs in~\eqref{eq:34} since
$\sigma = \Delta t/P\to0$ in that limit. We should stress, however,
that this limit is not reached in practice since the scheme
\eqref{eq:34discrete2} is used at small but finite $\Delta t$. We analyze next what happens when we 
adjust the size of $\sigma$  by
changing $\Delta t$ and/or the batch size~$P$.

\subsection{Dean's equation for particles with correlated noise}
\label{sec:deancorr} Lemma~\ref{th:lem1} indicates that we can
analyze the properties of~\eqref{eq:34trainz} instead of those
of~\eqref{eq:34discrete2}. 
To this end, we derive an equation for the empirical distribution $\mu^{(n)}_t$ in \eqref{eq:37} when $\{\thetab_i(t)\}_{i=1}^n$ satisfy the SDE~\eqref{eq:34trainz}; this calculation is operationally similar to the derivation of~\eqref{eq:38} but takes into account the extra drift term and the noise term in~\eqref{eq:34trainz}.  By applying It\^{o}'s formula to~\eqref{eq:85} we deduce
\begin{equation}
  \label{eq:42sgd}
  \begin{aligned}
    d \int_D \chi(\thetab) \mu^{(n)}_t(d\thetab)  & = \frac1n \sum_{i=1}^n
    \nabla \chi(\thetab_i(t)) \cdot d\thetab_i(t)  \\
    &+ \frac{\sigma}{2n} \sum_{i=1}^n \nabla
    \nabla\chi(\thetab_i(t)) : A([f-f_t^{(n)}],\thetab_{i}(t), \thetab_{i}(t) )dt
  \end{aligned}
\end{equation}
where $f_t^{(n)} = n^{-1} \sum_{i=1}^n \varphi(\cdot, \thetab_i(t)) = \int_D \varphi(\cdot, \thetab)\mu_t^{(n)}(d\thetab)$. Using ~\eqref{eq:34trainz} and the definition of $\mu^{(n)}_t$, this relation can be written as
\begin{equation}
  \label{eq:42sgd2}
  \begin{aligned}
    d \int_D \chi(\thetab) \mu^{(n)}_t(d\thetab)  & = \int_D 
    \nabla \chi(\thetab) \cdot \nabla V(\thetab,[\mu^{(n)}_t])\mu^{(n)}_t(d\thetab)dt\\
    & + \frac{\sigma}{2}\int_D \nabla \nabla\chi(\thetab) : A([f-f_t^{(n)}],\thetab, \thetab ) \mu^{(n)}_t(d\thetab)   dt\\
    &+ \frac{\sqrt{\sigma}}n \sum_{i=1}^n
    \nabla \chi(\thetab_i(t)) \cdot d\Bb_i(t) 
  \end{aligned}
\end{equation}
The drift terms in this equation are expressed in term of $\mu^{(n)}_t$; for the noise term, notice that its quadratic variation is
\begin{equation}
    \begin{aligned}
    &\Big\< \frac{\sqrt{\sigma}}n \sum_{i=1}^n
    \nabla \chi(\thetab_i(t)) \cdot d\Bb_i(t), \frac{\sqrt{\sigma}}n \sum_{i=1}^n
    \nabla \chi(\thetab_i(t)) \cdot d\Bb_i(t)\Big\>\\
    & = \sigma
    \int_{D\times D} 
    \nabla\chi(\thetab) \nabla\chi(\thetab') : A([f-f_t^{(n)}],\thetab, \thetab' )\mu^{(n)}_t(d\thetab)   
    \mu^{(n)}_t(d\thetab') dt 
    \end{aligned}
\end{equation}
This means that, in law, \eqref{eq:42sgd2} is equivalent to
\begin{equation}
  \label{eq:38trainz}
  \begin{aligned}
    d \int_D \chi(\thetab) \mu^{(n)}_t(d\thetab)  & = \int_D 
    \nabla \chi(\thetab) \cdot \nabla V(\thetab,[\mu^{(n)}_t])\mu^{(n)}_t(d\thetab)dt\\
    & + \frac{\sigma}{2} 
    \int_D \nabla \nabla\chi(\thetab) : A([f-f_t^{(n)}],\thetab, \thetab ) \mu^{(n)}_t(d\thetab)dt   \\
    &+ \sqrt{\sigma} \int_D 
    \nabla \chi(\thetab) \cdot d\etab^{(n)}_t(d\thetab) 
  \end{aligned}
\end{equation}
where $d\etab^{(n)}_t(d\thetab) $ is vector-valued random measure, white in time, and with quadratic variation
\begin{equation}
\label{eq:quadeta}
    \< d\etab^{(n)}_t(d\thetab), d\etab^{(n)}_t(d\thetab')\> = A([f-f_t^{(n)}],\thetab, \thetab' )
    \mu^{(n)}_t(d\thetab)  
    \mu^{(n)}_t(d\thetab') dt
\end{equation}

The first term at the right hand side of~\eqref{eq:38trainz} is
the same as in the weak form of~\eqref{eq:38}. This is because these terms come
from the drift terms in~\eqref{eq:34trainz}, which also coincide with
those in~\eqref{eq:34}. However, \eqref{eq:38trainz} also contains
additional terms that were absent
in~\eqref{eq:38}---note that these terms are different from those in the standard Dean's equation, because the noise term in~\eqref{eq:34trainz} is correlated between the particles, instead of being independent.

\subsection{LLN for SGD}
\label{sec:llntrain}

If we want the result established in Proposition~\ref{th:lln} to apply and also for the approximation error to vanish as $n\to\infty$, we need to make the additional terms in~\eqref{eq:38trainz} compared to~\eqref{eq:38} higher order. This can be done by scaling $\sigma$
with some inverse power of~$n$. Specifically, we will
set
\begin{equation}
  \label{eq:83}
  \sigma = a n^{-2\alpha}\quad  \text{for some} \quad  a>0\quad
  \text{and}\quad  \alpha>0
\end{equation}
This scaling can be achieved by choosing, e.g., $ P = O(n^{2\alpha})$, which amounts to increasing the batch size with $n$.
The choice~\eqref{eq:83} implies that the last two terms in~\eqref{eq:38trainz} 
disappear in the limit as
$n\to\infty$. Therefore,   we formally conclude that
$\mu^{(n)}_t\rightharpoonup\mu_t$ as $n\to\infty$, where $\mu_t$ solves the same
deterministic equation~\eqref{eq:39} as before. This implies that
$\lim_{n\to\infty} f^{(n)}_t =f_t= \int_{D} \varphi(\cdot,\thetab) \mu_t(d\thetab)
$ satisfies \eqref{eq:48} and is such that $f_t\to f $ as
$t\to\infty$. 
In particular, both the LLN and the global convergence result in Proposition~\ref{th:lln} still hold if the assumption in this proposition are met and we use the solution
of~\eqref{eq:38train2} in~\eqref{eq:68}.
In turn, we can also conclude that
this proposition holds up to discretization errors in $\Delta t$
if we use the solution of~\eqref{eq:34discrete2} in~\eqref{eq:68}. 
Importantly, the covariance associated with the estimator for the gradient,
defined in~\eqref{eq:161}, satisfies
\begin{equation}
  \label{eq:126}
  \forall (\thetab,\thetab')\in D\times D \ : \qquad \lim_{t\to\infty} A([f_t-f],\thetab,\thetab') = 0.
\end{equation}
This property will be useful later.

\subsection{CLT for SGD}
\label{sec:llnclt}

Turning our attention to the fluctuations of $\mu^{(n)}_t$ around $\mu_t$,
notice that there are two sources of them: some are intrinsic to the
discrete nature of the particles apparent in $\mu^{(n)}_t$, and scale
 as $O(n^{-1/2})$ for all $t<\infty$ and possibly as $O(n^{-\xi})$ for any
$\xi<1$ as $t\to\infty$, as discussed in Sec.~\ref{sec:first}. Other fluctuations
come from the noise term in~\eqref{eq:38trainz}, and scale as
$O(n^{-\alpha})$ when~\eqref{eq:83} holds. The It\^o drift terms
proportional to $\sigma = a n^{-2\alpha}$ in~\eqref{eq:38trainz}
always make higher order contributions.

We first consider $t<\infty$ and subsequently examine the limit $t\to\infty$ in Sec.~\ref{sec:SGDlvlt}. 
In the present case, we first observe that if $\alpha\ge\frac12$, then for all $t<\infty$ the fluctuations due to the noise
in~\eqref{eq:38trainz} are negligible compared to the intrinsic
ones from discreteness, and we are back to the GD situation studied in
Sec.~\ref{sec:weighted}.
In contrast, if $\alpha\in (0,\frac12)$, for all $t<\infty$  the fluctuations due to the noise in~\eqref{eq:38trainz}  dominate the intrinsic ones
from discreteness, so let us focus on this case from now on.  
To quantify these fluctuations, we can introduce
$ n^{\alpha} (\mu^{(n)}_t-\mu_t)$, write an equation for this scaled discrepancy,
and take the limit as $n\to\infty$.  
The derivation proceeds akin to the derivation of~\eqref{eq:38omlim} and leads to the conclusion
that, as $n\to\infty$, $n^{\alpha} (\mu^{(n)}_t-\mu_t) \rightharpoonup \omega^{(\alpha)}_t$ in law which
satisfies
\begin{equation}
  \label{eq:38trainzlim}
  \begin{aligned}
    d \int_D \chi(\thetab) \omega^{(\alpha)}_t(d\thetab)  & = \int_D 
    \nabla \chi(\thetab) \cdot  \nabla V(\thetab,[\mu_t])\omega^{(\alpha)}_t(d\thetab) dt \\
    & + \int_D 
    \nabla \chi(\thetab) \cdot  \nabla F(\thetab,[\omega^{(\alpha)}_t]) \mu_t(d\thetab) dt\\
    & + \sqrt{a} \int_D 
    \nabla \chi(\thetab) \cdot d\etab_t(d\thetab) 
  \end{aligned}
\end{equation}
in which $d\etab_t(d\thetab) $ is vector valued random measure, white in time, and with quadratic variation (compare~\eqref{eq:quadeta})
\begin{equation}
\label{eq:quadetalim}
    \< d\etab_t(d\thetab), d\etab_t(d\thetab')\> = A([f-f_t],\thetab, \thetab' )
    \mu_t(d\thetab)  
    \mu_t(d\thetab') dt
\end{equation}
Equation~\eqref{eq:38trainzlim} should be solved with zero initial condition, since the $O(n^{-1/2})$ fluctuations arising from the initial condition are higher order compared to scaling $O(n^{-\alpha})$ we picked to obtain~\eqref{eq:38trainzlim}. Since~\eqref{eq:38trainzlim} is linear in $\omega^{(\alpha)}_t$ with additive noise, it indicates that $\omega^{(\alpha)}_t$ a Gaussian process with mean zero and thereby fully characterized by its covariance (we omit the equation for brevity).  
This also implies that
\begin{equation}
    n^{\alpha} (f^{(n)}_t - f) \to g^{(\alpha)}_t \qquad \text{in law as $n\to\infty$} 
\end{equation}
where $g^{(\alpha)}_t $ is a Gaussian process whose evolution equation (cf. the derivation of ~\eqref{eq:75}) gives,
\begin{equation}
  \label{eq:38train2}
  \begin{aligned}
    d g^{(\alpha)}_t & = - \int_\Omega M([\omega^{\alpha}_t],\xb,\xb') \left(f_t(\xb)
      -f(\xb')\right) \nu(d\xb') dt \\
      & \quad-\int_\Omega M([\mu_t],\xb,\xb')
    g^{(\alpha)}_t(\xb') \nu(d\xb') dt  + \sqrt{a}\,  d\!\zeta_t(\xb)
  \end{aligned}
\end{equation}
where $M([\mu],\xb,\xb')$ is given in~\eqref{eq:96}, and the quadratic
variation of $d\zeta_t$ is that of $\int_{D} \varphi(\cdot,\thetab) d\eta_t(d\thetab)$.
Explicitly, 
\begin{equation}
  \label{eq:60}
  \begin{aligned}
    &\< d\zeta_t(\xb),d\zeta_t(\xb')\>\\ 
    &= \int_{\Omega}
    M([\mu_t],\xb,\xb'')M([\mu_t],\xb',\xb'')
    \left|f_t(\xb'')-f(\xb'')\right|^2 d\nu(\xb'') dt\\
    & -\int_{\Omega} M([\mu_t],\xb,\xb'') \left(f_t(\xb'')-f(\xb'')\right) d\nu(\xb'')\\
    & \times \int_{\Omega} M([\mu_t],\xb',\xb'') \left(f_t(\xb'')-f(\xb'')\right) d\nu(\xb'')dt.
  \end{aligned}
\end{equation}
The SDE \eqref{eq:38train2} should be solved with zero initial condition, $g_0^{(\alpha)}=0$. Since it is linear in $g^{(\alpha)}_t$ with additive noise, it defines a Gaussian process with mean zero and is specified by its covariance
\begin{equation}
    C^{(\alpha)}_t(\xb,\xb') = \EE [ g^{(\alpha)}_t(\xb) g^{(\alpha)}_t(\xb)]
\end{equation}
where $\EE$ denotes expectation over the noise $d\zeta_t$ (that is, over the data in the batches used in SGD).  With this calculation, we have established

\begin{proposition}[CLT for SGD]
\label{th:cltsgd1}
  Consider 
  \begin{equation}
  \label{eq:flucta}
      g^{(\alpha,n)}_t = n^{\alpha-1} \sum_{i=1}^n \left(\varphi(\cdot, \thetab_i(t)) - f_t\right) = n^{\alpha}  \big(f^{(n)}_t - f_t\big)
  \end{equation}
  with $\{\thetab_i(t)\}_{i=1}^n$ solution to the SDE~\eqref{eq:34trainz} with $\sigma = an^{-2\alpha}$, $\alpha \in (0,\frac12)$, and $f_t$ solution to~\eqref{eq:48}. Then, as $n\to\infty$, $g^{(\alpha,n)}_t$ converges in law towards the Gaussian process $g^{(\alpha)}_t$ solution of~\eqref{eq:38train2} for $g^{(\alpha)}_0=0$.
\end{proposition}

\subsection{Fluctuations in SGD at long and very long times}
\label{sec:SGDlvlt}

The noise $d\zeta_t$ in~\eqref{eq:38train2}
has the remarkable property that it
self-quenches as $t\to\infty$ if the conditions of Proposition~\ref{th:lln} are met and $f_t\to f$ as $t\to\infty$ and therefore, from~\eqref{eq:60}:
\begin{equation}
    \forall \xb,\xb'\in \Omega \ : \quad \lim_{t\to\infty}\< d\zeta_t(\xb),d\zeta_t(\xb')\> =0.
\end{equation}
Since the first drift term in~\eqref{eq:38train2} also goes to zero when $f_t\to f$ and the second drift term is a damping term because $M([\mu_t],\xb,\xb')$ is positive definite for all $t<\infty$, we know that $g^{(\alpha)}_t$ will be controlled as $t\to\infty$, i.e. $C^{(\alpha)}_t$ as a limit. In addition, if $\supp \hat \mu^* = \hat D$ where $\hat \mu^* = \int_\RR \mu^*(dc,\cdot) = \lim_{t\to\infty} \int_\RR \mu^*(dc,\cdot)$, then $M([\mu_t],\xb,\xb')$ is positive definite for all $t<0$ and in the limit as $t\to\infty$, and the solution to~\eqref{eq:38train2} goes to zero. Using the definition the Gaussian process $g^{(n,\alpha)}_t$ defined in~\eqref{eq:flucta}, we can summarize this result into:

\begin{proposition}[Fluctuations in SGD at long time]
  \label{th:train}
  Under the conditions of Proposition~\ref{th:lln}, if $f_t^{(n)}$ is given by by~\eqref{eq:68} with
  $\{\thetab_i(t)\}_{i=1}^n$ solution of~\eqref{eq:34trainz} with $\sigma = an^{-2\alpha}$, $\alpha \in (0,\frac12)$, and initial
  condition drawn from $\PP_{\text{in}}$, and $f_t$ solves~\eqref{eq:48}, then 
  \begin{equation}
      \lim_{t\to\infty} \lim_{n\to\infty} n^{2\alpha} \EE\int_\Omega |f^{(n)}_t(\xb) - f_t(\xb)|^2 \nu(d\xb) 
      = \lim_{t\to\infty} C^{(\alpha)}_t(\xb,\xb') \quad \text{exists}
  \end{equation}
  In addition, if $\supp \hat \mu^* = \hat D$, then this limit is zero.
\end{proposition}

If $\supp \hat \mu^* = \hat D$ where $\hat \mu^* = \int_\RR \mu^*(dc,\cdot) = \lim_{t\to\infty} \int_\RR \mu^*(dc,\cdot)$, then $M([\mu_t],\xb,\xb')$ is positive definite for all $t<0$ and in the limit as $t\to\infty$. In that case, the only
fixed point of~\eqref{eq:38train2} is zero. Since in this case we also know that the fluctuations from the initial conditions disappear on scale $O(n^{\-\xi})$ for any $\xi<0$, we can proceed as in Sec.~\ref{sec:first} and adjust $\alpha$ all the way up to 1 instead of $\frac12$. That is, we can generalize Proposition~\ref{th:cltgvlt} into

\begin{proposition}[Fluctuations in SGD at very long times]
  \label{th:train2}
  Under the conditions of Proposition~\ref{th:train}, if $\supp \hat \mu^* = \hat D$, then for any $\alpha\in(0,1)$,
  \begin{equation}
    \label{eq:147}
     \lim_{n\to\infty} n^{2\alpha} \EE\int_\Omega |f^{(n)}_{a_n}(\xb) - f(\xb)|^2 \nu(d\xb) 
      = 0
  \end{equation}
if $a_n$ grows with $n$ and is such that $\lim_{n\to\infty} a_n/\log n = \infty$---here $\EE$ denotes expectation over both the initial condition, $\PP_{\text{in}}$, and the noise in~\eqref{eq:34trainz}.  
\end{proposition}

\section{Illustrative example: 3-spin model on the high-dimensional
  sphere}
\label{sec:examples}

To test our results, we use a function known for its complex
features in high-dimensions: the spherical 3-spin model,
$f: S^{d-1}(\sqrt{d}) \to \RR$, given by
\begin{equation}
  \label{eq:psin}
  f(\xb) = \frac1d \sum_{p,q,r=1}^d a_{p,q,r} x_p x_q x_r,  \qquad \xb
  \in S^{d-1}(\sqrt{d}) \subset \RR^d
\end{equation}
where the coefficients $\{a_{p,q,r}\}_{p,q,r=1}^d$ are independent
Gaussian random variables with mean zero and variance one. The
function~\eqref{eq:psin} is known to have a number of critical points
that grows exponentially with the dimensionality
$d$~\cite{Auffinger:2013kq,Sagun:2014tg,Auffinger:2012gh}. We note
that previous works have sought to draw a parallel between the glassy
3-spin function and generic loss
functions~\cite{Choromanska:2014ui}, but we are not exploring such an
analogy here. Rather, we simply use the function~\eqref{eq:psin} as a
difficult target for approximation by neural networks. That is,
throughout this section, we train networks to learn $f$ with a
particular realization of $a_{p,q,r}$ and study the accuracy of that
representation as a function of the number of particles $n$.

\subsection{Learning with Gaussian kernels}
\label{sec:gauss}

We first consider the case when $D = S^{d-1}(\sqrt{d})$ and we use
\begin{equation}
  \label{eq:90}
  \varphi(\xb,\zb) = e^{-\tfrac12 \alpha |\xb-\zb|^2}
\end{equation}
for some fixed $\alpha >0$. In this case, the parameters
are elements of the domain of the function (here the $d$-dimensional
sphere). 
Note that, since $|\xb|=|\zb| = \sqrt{d}$, up to an
irrelevant constant that can be absorbed in the weights $c$, we can
also write~\eqref{eq:90} as
\begin{equation}
  \label{eq:90red}
  \varphi(\xb,\zb) = e^{-\alpha \xb\cdot \zb}
\end{equation}
This setting allow us to simplify the problem. Using
\begin{equation}
  \label{eq:119}
  f^{(n)}(\xb) = \frac1n \sum_{i=1}^n c_i \varphi(\xb,\zb_i)
  = \frac1n \sum_{i=1}^n c_i e^{-\alpha \xb\cdot \zb_i},
\end{equation}
we can use as alternative loss
\begin{equation}
\label{eq:otherloss}
    \mathcal{L}[f^{(n)}] = -\frac1n \sum_{i=1}^n c_i f(\zb_i) + \frac1{2n^2} \sum_{i,j=1}^n c_i c_j \varphi(\zb_i,\zb_j)
\end{equation}
i.e. eliminate the need for data beside the set $\{\zb_i\}_{i=1}^n$. In terms of the empirical distribution, the loss can be represented as
\begin{equation}
    \begin{aligned}
    \mathcal{L}[f^{(n)}] &= -\int_{\hat D} f(\zb) \gamma^{(n)}(d\zb) + \frac1{2} \int_{\hat D\times\hat D}
    \varphi(\zb,\zb')\gamma^{(n)}(d\zb)\gamma^{(n)}(d\zb')
    \end{aligned}
\end{equation}
where $\gamma^{(n)}=\int_\RR c \mu^{(n)}(dc,\cdot)$. Viewed as an integral kernel, $\varphi$ is positive definite, as a result the loss is a convex functional of $\gamma^{(n)}$ (or $\mu^{(n)}$). Hence, the results established above apply to this special case, as well. 
The GD flow on the loss~\eqref{eq:otherloss} can now be written explicitly as
\begin{equation}
  \label{eq:34RBFpspin}
  \left\{
    \begin{aligned}
      \dot\zb_i &= c_i \nabla f(\zb_i) +
      \frac{\alpha}n\sum_{j=1}^n
      c_i c_j \zb_j e^{-\alpha \zb_i\cdot \zb_j} -\lambda_i \zb_i \\
      \dot c_i &= f(\zb_i) - \frac1n\sum_{j=1}^n c_j e^{-\alpha
        \zb_i\cdot \zb_j}
    \end{aligned}
  \right.
\end{equation}
where $ -\lambda_i \zb_i $ is a Lagrange multiplier term added to
enforce $|\zb_i|= \sqrt{d}$ for all $i=1,\ldots, n$, $f(\xb)$ is
given by~\eqref{eq:psin}, and $\nabla f(\zb)$ is given componentwise by
\begin{equation}
  \label{eq:132}
  \frac{\partial f}{\partial z_p} = \frac1d\sum_{q,r=1}^d
  \left(a_{p,q,r} + a_{r,p,q}+ a_{q,r,p}\right) z_q z_r,
\end{equation}
As is apparent from~\eqref{eq:34RBFpspin} the advantage of using radial
basis function networks (or, in fact, any unit $\hat \phi$ which is (i) such that $\hat D = \Omega$ and (ii) positive definite) is that we can use $f(\xb)$ and the unit
$\varphi(\xb,\zb)$ directly, and do not need to evaluate $\hat F(\zb)$ and
$\hat K(\zb,\zb')$ (that is, we need no batch). In other words, the cost of
running~\eqref{eq:34RBFpspin} scales like $(dn)^2$, instead of $P(Nn)^2$ in
the case of a general network optimized by SGD with a batch of size
$P$ and $\zb\in \hat D \subset \RR^N$. If we make $P$ scale with $n$, like
$P=C n^{2\alpha}$ for some $C>0$, as we need to do to obtain the
scalings discussed in Sec.~\ref{sec:stochgrad}, the cost of SGD
becomes $N^2n^{2+2\alpha}$, which is quickly becomes much worse than
$(dn)^2$ as $n$ grows.

\begin{figure}[t]
\includegraphics[width=0.45\linewidth]{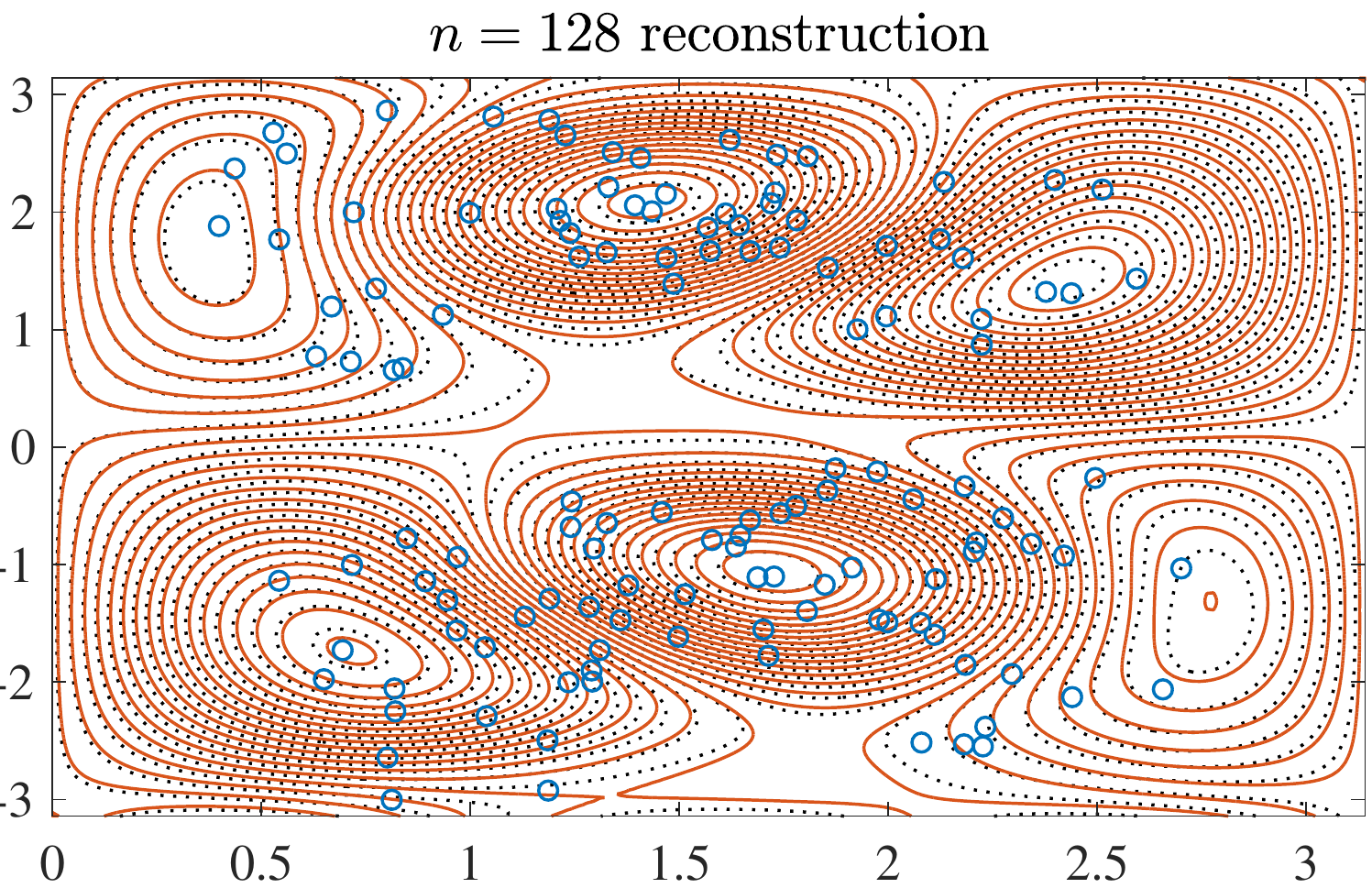}
\includegraphics[width=0.45\linewidth]{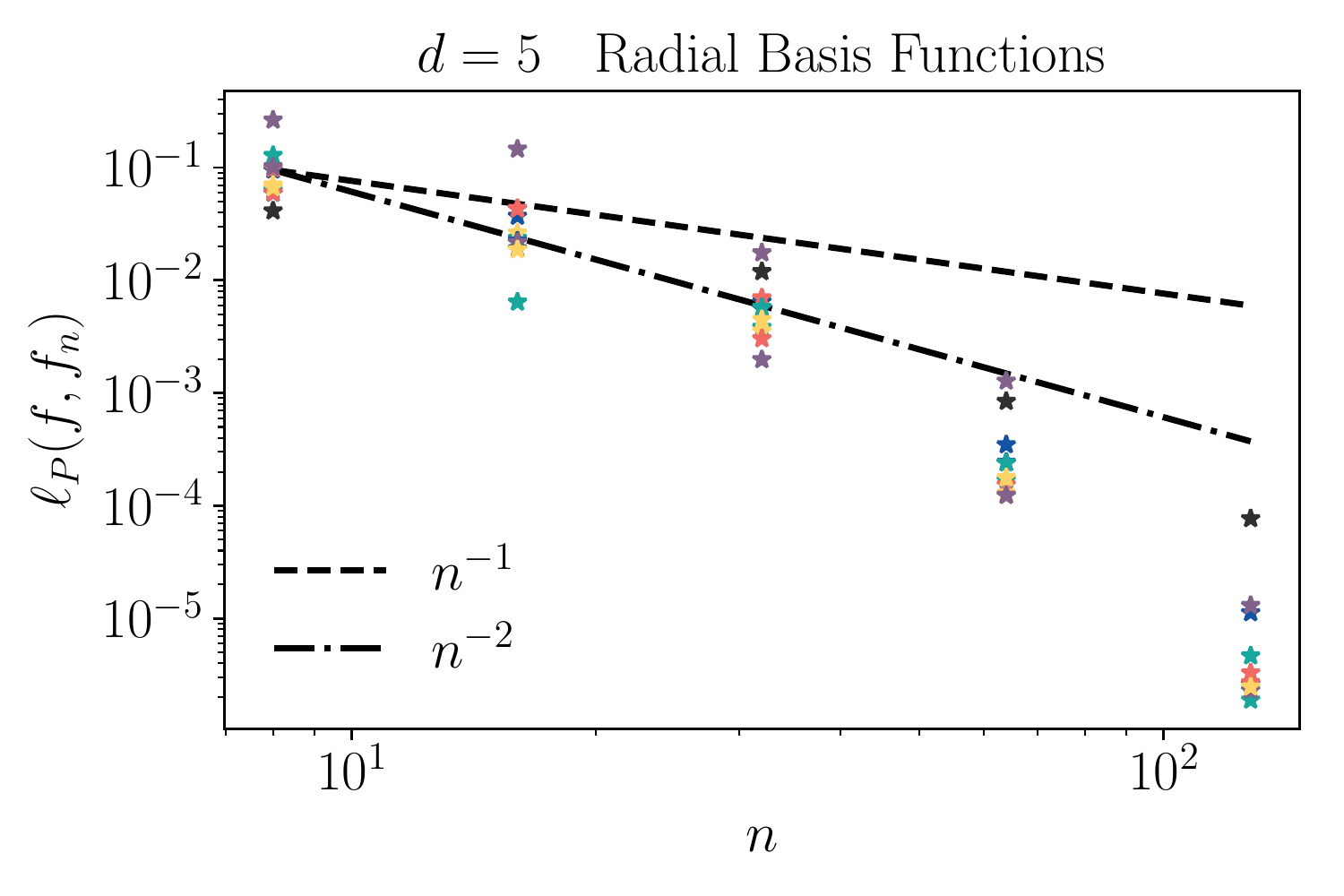}

\caption{Left panel: Comparison between the level sets of the original
  function $f$ in~\eqref{eq:psin} (black dotted curves) and its approximation
  by the neural network in~\eqref{eq:119} with $n= 128$ and $d=5$ in
  the slice defined by~\eqref{eq:slice}. Also shown are the projection
  in the slice of the particle position. Right panel: empirical loss in
  ~\eqref{eq:lossempirical} vs $n$ at the end of the calculation. The
  stars show the empirical loss for 10 independent
  realizations of the coefficients $a_{p,q,r}$ in~\eqref{eq:psin}.  }
\label{fig:5DRBF}
\end{figure}

We tested the representation~\eqref{eq:119} in $d=5$ using $n=16$, 32,
64, 128, and 256 and setting $\alpha = 5/d= 1$. The training was done
by running a time-discretized version of~\eqref{eq:34RBFpspin} with
time step $\Delta t = 10^{3}$ for $2 \times 10^5$ steps: during the
first $10^5$ we added thermal noise to
\eqref{eq:34RBFpspin}, which we then remove during the second half of
the run. The representation~\eqref{eq:119} proves to be accurate even
at rather low value of $n$: for example, the right panel of
Fig.~\ref{fig:5DRBF} shows a contour plot of the original function $f$
and its representation $f^{(n)}$ with $n=128$ through a slice of the
sphere defined as
\begin{equation}
  \label{eq:slice}
  \xb(\theta) = \sqrt{d} \left(\sin(\theta)\cos(\phi),
    \sin(\theta)\sin(\phi),
    \cos(\theta), 0,0\right), 
\end{equation}
with $\theta\in[0,\pi]$ and $\phi\in [0,2\pi)$.
The level sets of both functions are in good agreement. Also shown on
this figure is the projection on the slice of the position of the 64
particles on the sphere. In this result, the parameters $c_i$ take
values that are initially uniformly distributed by about
$-40 d^2 = -10^3$ and $40 d^2 = 10^3$. To test the accuracy of the
representation, we used the following Monte Carlo estimate of the loss
function
 \begin{equation}
   \label{eq:lossempirical}
   \mathcal{L}_P[f^{(n)}_t] = \frac{1}{2P} \sum_{p=1}^P \left|f(\xb_p) - f^{(n)}_t(\xb_p)\right|^2.
 \end{equation}
 This empirical loss function was computed with a batch of $10^6$
 points $\xb_p$ uniformly distributed on the sphere. The value
 \eqref{eq:lossempirical} calculated at the end of the calculation is
 shown as a function of $n$ in the right panel of
 Fig.~\ref{fig:5DRBF}: the empty circles show~\eqref{eq:lossempirical}
 for 4 individual realizations of the coefficient $a_{p,q,r}$
 in~\eqref{eq:psin}, the full circle shows the average
 of~\eqref{eq:lossempirical} over these 4 realizations. The blue line
 scale as $n^{-1}$, the red one as $n^{-2}$: as can be seen, the
 empirical loss decays with $n$ faster than $n^{-1}$, which is as
 expected.

\subsection{Learning with single layer networks with sigmoid
  nonlinearity}
\label{sec:sigmo}

\begin{figure}[t]
  \includegraphics[width=0.9\linewidth]{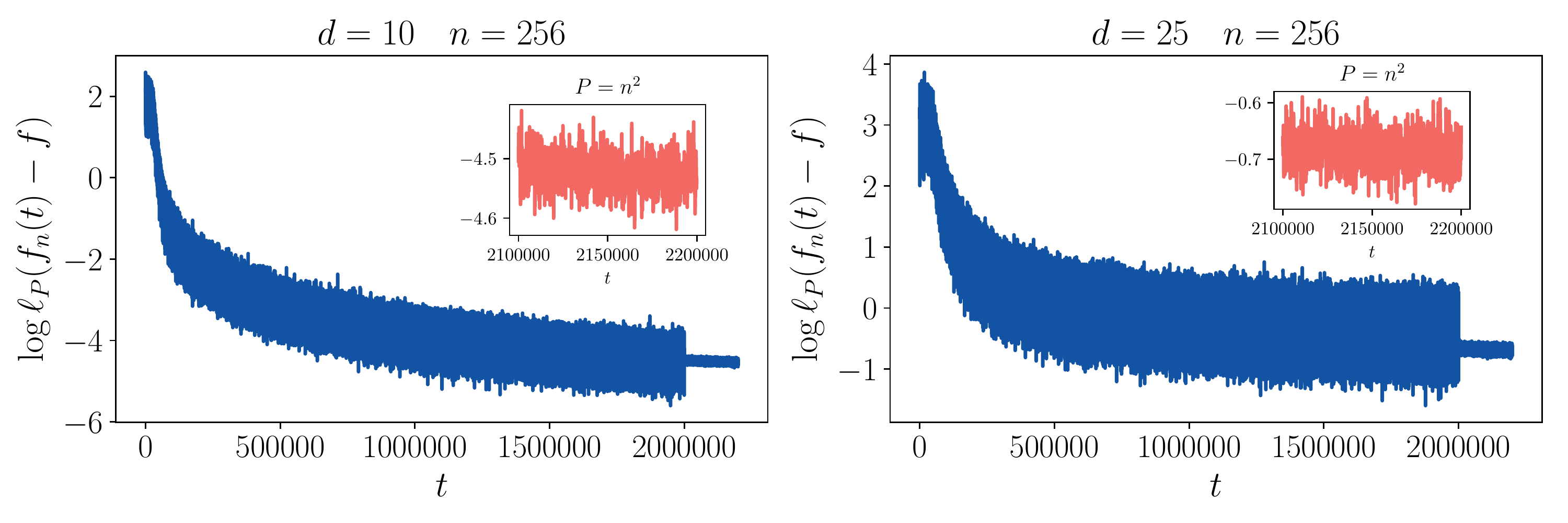}
\caption{The $\log$ of the empirical loss in~\eqref{eq:lossempirical}
  as a function of training time by SGD for the sigmoid neural network
  in $d=10$ (left panel) and $d=25$ (right panel). At time $t=2\times 10^6$, the
  batch size is increased to initiate a quench. The insets show the
  $\log$ of the empirical loss as a function of time during the final
  $10^5$ time steps of training. }
\label{fig:quench}
\end{figure}

To further test our predictions and also assess the learnability of
high dimensional functions, we used $3$-spin models in $d=10$ and 25
dimensions, which we approximated with a single-layer neural network
with sigmoid nonlinearity parameterized by
$\zb=(\ab,b) \in D= \RR^{d+1}$, with $\ab \in \RR^{d}$, $b\in \RR$,
and
\begin{equation}
  \label{eq:133}
  \varphi(\xb,\zb) = h(\ab\cdot \xb + b).
\end{equation}
This gives
\begin{equation}
  \label{eq:fnsigmo}
  f^{(n)}(\xb) = \frac1n \sum_{i=1}^n c_i h(\ab_i \cdot \xb + b_i)
\end{equation}
where $h(z) = 1/(1+e^{-z})$.  Simple networks like these, as opposed
to deep neural with many parameters, provide greater assurance that we
have trained sufficiently to test the scaling.

\begin{figure}[t]
\includegraphics[width=0.9\linewidth]{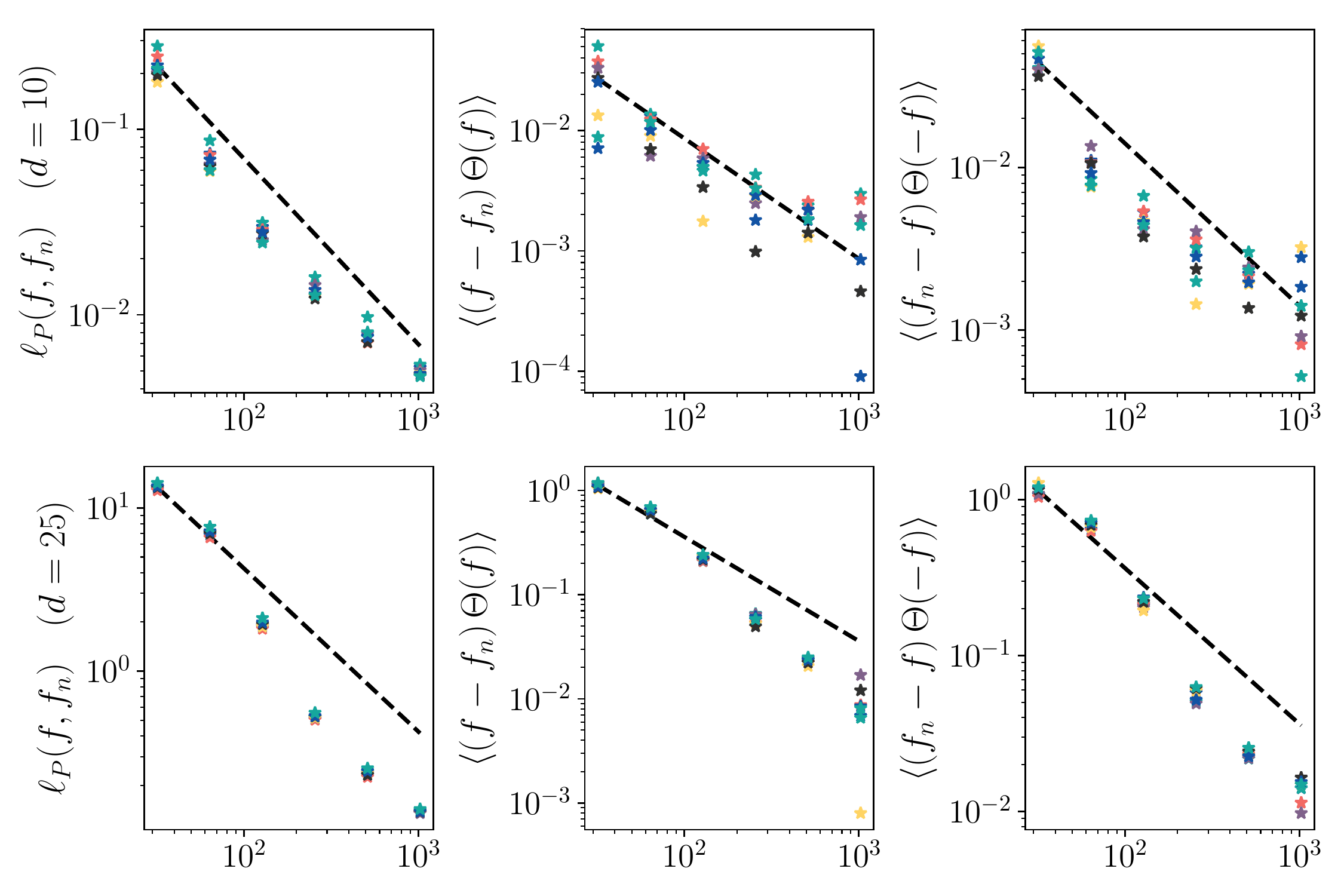}
\caption{Error scaling for single layer neural network with sigmoid
  nonlinearities. Upper row: $d=10$; lower row: $d=25$. The first
  column shows the empirical loss in~\eqref{eq:lossempirical}, the
  second column shows~\eqref{eq:test2}, and the third column
  shows~\eqref{eq:test2} with $\Theta(f)$ replaced by
  $\Theta(-f)$. The stars show the results for 10 different
  realizations of the coefficients $a_{p,q,r}$ in~\eqref{eq:psin}: the
  dashed lines decay as $n^{-1}$, consistent with the predictions
  in~\eqref{eq:147} and~\ref{th:cltg} .}
\label{fig:scaling}
\end{figure}

We trained the model in~\eqref{eq:fnsigmo} using SGD with an initial
batch size of $P = \lfloor n/5\rfloor$ points uniformly sampled on the
sphere for $2\times 10^6$ time steps, resampling a new batch at every
time step: this corresponds to choosing $\alpha = 1/2$ in the notation
of Sec.~\ref{sec:stochgrad}.  Towards the end of the trajectory, we
initiated a partial quench by increasing the batch size to
$P=\lfloor (n/5)^2\rfloor$ (i.e $\alpha = 1$) which we run for an
additional $2\times 10^5$ time steps.  Fig.~\ref{fig:quench} shows the
empirical loss in~\eqref{eq:lossempirical} calculated over the batch
as a function of training time during the optimization with $n=256$
particles and $d=10$ (left panel) and $d=25$ (right panel). Note that
the lack of intermediate plateaus in the loss during training is
consistent with our conclusion that the dynamics effectively descends
on a quadratic energy landscape (i.e. the loss function itself) at the
level of the empirical distribution of the particles.  After the
quench the empirical loss shows substantially smaller fluctuations as
a function of time which helps to reduce the fluctuating error.  The
inset shows the final $10^5$ time steps in which there is negligible
downward drift, indicating convergence towards stationarity at this
batch size.

In these higher dimensional examples, we tested the scaling with three
different observables.  First, we considered the empirical loss
function in~\eqref{eq:lossempirical} which we computed over a batch of
size $\hat P=10^5$ larger than $P$. As shown in the two right panels
Fig.~\ref{fig:scaling}, $\mathcal{L}_{\hat P}[f^{(n)}_t]$ scales as $n^{-1}$, as expected. 
We
also tested the estimate in~\eqref{eq:147} using
\begin{equation}
  \label{eq:test2}
  \frac{1}{\hat P} \sum_{p=1}^{\hat P}
  \Theta\left(f(\xb_p)\right)\left(f(\xb_p)- f^{(n)}_t(\xb_p)\right),
\end{equation}
and similarly with $\Theta\left(-f(\xb_p)\right)$: here $\Theta$
denotes the Heaviside function.  The result is shown in the four right
panels in Fig.~\ref{fig:scaling}: \eqref{eq:test2} scales as $n^{-1}$,
consistent with~\eqref{eq:147} and our choice of $\alpha =1$.

\begin{figure}[t]
\includegraphics[width=0.9\linewidth]{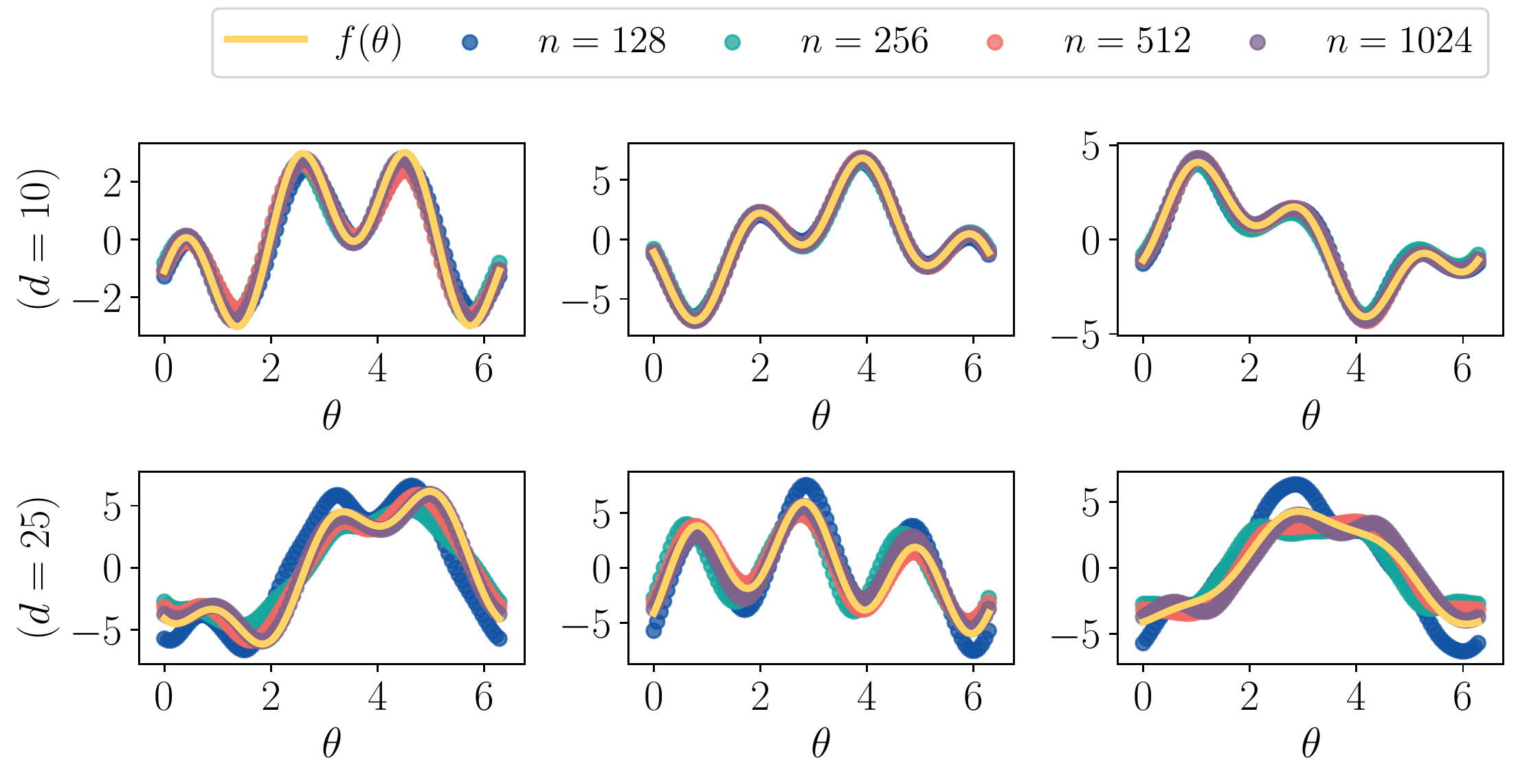}
\caption{One dimensional slices through the $d=10$ (upper row) and
  $d=25$ (lower row) neural net representation $f^{(n)}$ are shown below a
  yellow curve with the target function $f$. In $d=10$, the function
  representations clearly capture the main features of the target
  function, with only small scale deviations. In $d=25$ there is
  remarkably good signal when $n=1024$ while the smaller neural
  network is less able to faithfully represent the target function.  }
\label{fig:slices}
\end{figure}

To provide further confidence in the quality of the representations,
we also made a visual comparison by plotting $f$ and $f^{(n)}$ along great
circles of the sphere.  We do so by picking $i\neq j$ in
$\{1,\cdots, d\}$ and setting
$\xb = \xb(\theta) = (x_1(\theta), \dots x_d(\theta))$ with
\begin{equation}
  x_i(\theta) = \sqrt{d}\cos(\theta), \qquad x_j(\theta) = \sqrt{d}
  \sin(\theta),
  \qquad
  x_k(\theta) = 0 \quad \forall k\not=i,j.
\end{equation}
In Fig.~\ref{fig:slices} we plot $f(\xb(\theta)$ and
$f^{(n)}(\xb(\theta))$ along three great circles for $d=10$ and $d=25$. As
can be seen, the agreement is quite good and confirms the quality of
the final fit.  A strong signal is present in $d=25$ with $n=1024$, a
remarkable fact when considering that if we had only two grid points
per dimension, the total number of points in the grid would be
$2^{25} = 33,554,432$.

\section{Concluding remarks}
\label{sec:conclu}

Viewing parameters as particles with the loss function as interaction
potential enables us to leverage a powerful theoretical apparatus
developed to analyze problems from statistical physics.
Using these ideas, we can analyze the approximation quality and the
trainability of neural network representations of high-dimensional
functions.  Several insights emerge from our analysis based on this
viewpoint: First, these tools show the dynamical realizability of the Universal Approximation
Theorems, a direct consequence of the Law of Large Numbers for the empirical
distribution of the parameters.  
Specifically, we conclude that
the empirical distribution effectively descends on the quadratic loss
function landscape when the number $n$ of parameters in the network is
large. This confirms the empirical observation that wide neural networks
are trainable despite the non-convexity of the loss function viewed
from the individual particles perspective (as opposed to that of their
empirical distribution). Secondly, we have derived a Central Limit
Theorem for the empirical distribution of the parameters, specifying
the approximation error of the neural network representation and showing
that it is universal.

We derived these results first in the context of gradient descent
dynamics; however, our conclusions also apply to
stochastic gradient descent. The analysis indicates how the parameters
in SGD should be chosen, in particular how the batch size should be
scaled with $n$ given the time step used in the scheme, which can be
done towards the end of training.

These results were derived for a quadratic loss, $\mathcal{L}[f^{(n)}]= \tfrac12 \EE_\nu|f-f^{(n)}|^2$. However, they do generalize to other losses  as long as they are convex in~$f^{(n)}$.

We also worked in the limit of an infinite amount of training data, an idealized setting that does not address the error incurred from a finite data set.
For a neural network trained on a  dataset of $P$ points, $\{\xb_p\}_{p=1}^P$, we can decompose the ``generalization'' error into components that involve the approximation error and the error from the finiteness of the data,
%Finally we stress that there is another important aspects of the problem we have not addressed here, namely the test error, i.e. the error on the exact loss when the network is trained on a dataset of fixed size. The training error we obtained here is just one component of this test error, though an important one since, given a data set $\{\xb_p\}_{p=1}^P$, we can always decompose
\begin{equation}
    \label{eq:testerror}
    \EE_\nu |f-f_P^{(n)}|^2 \le  \EE_\nu |f-f_P|^2+  \EE_\nu |f_P-f_P^{(n)}|^2
\end{equation}
where $f_P$ and $f^{(n)}_P$ are the approximations of $f$ we can get if we train the network on the empirical loss build on $\{\xb_p\}_{p=1}^P$ with finitely ($n<\infty$) or infinitely ($n\to\infty$) many units, respectively.  Our results give direct insight on the second term at the right hand side  of
\eqref{eq:testerror}. 
We leave assessments of the first term for future work.
%We will investigate elsewhere~\cite{brurotvde:neurips19} how the approach proposed here can also be used to estimate the first term after training.

Our numerical results not only confirm our predictions, they emphasize the
capability of neural networks to represent high-dimensional function
accurately with a relatively modest number of adjustable
parameters. Needless to say, the computational achievements of neural networks open the door to developments
in scientific computing that we are only beginning to grasp.  Such
applications may benefit from better understanding how the specific
architecture of the neural networks affects the approximation error
and trainability, not in the general terms of their scaling with $n$
that we analyzed here, but in the details of the constant involved.

\appendix

\section{Training at finite (but small) temperature}
\label{sec:finiteT}

For completeness, let us consider here the case when noise-terms are
added in~\eqref{eq:34} and the ODEs become stochastic
differential equations (SDEs).  Additive noise
addresses the non-uniqueness issues encountered in
Sec.~\ref{sec:weighted}. To formulate the resulting SDEs, we
need a distribution $\mu_0\in \mathcal{M}_+(D)$, used to regularize
the dynamics. We specify its properties via:

\begin{assumption}
  \label{as:meas}
  The distribution $\mu_0$ (i) has a density $\rho_0$ that is continuously 
  differentiable, $\rho_0\in C^1(D)$; (ii) is such that  $\supp (\mu_0) = D$; and (iii) satisfies 
  \begin{equation}
    \label{eq:101}
    \forall b \in \RR \ \ : \ \  \int_{
      \RR} e^{b c} \mu_0(dc,\cdot) <\infty \quad \text{and} \quad \int_\RR c \mu_0(dc,\cdot) = 0.
  \end{equation}
\end{assumption}
We then replace~\eqref{eq:34} with the SDEs
\begin{equation}
  \label{eq:34fT}
  \begin{aligned}
      d\thetab_i &=  \nabla F(\thetab_i) dt -
      \frac1n\sum_{j=1}^n \nabla K(\thetab_i,\thetab_j)
      dt \\
      & +  (\beta n)^{-1}
      \nabla \log \rho_0(\thetab_i) dt+
      \sqrt{2}(\beta n)^{-1/2} d\Wb_i,
      \end{aligned}
\end{equation}
for $i=1,\ldots, n$.  Here $\Wb_i$ are $n$ independent
Wiener processes, taking values in $D$, 
and $\beta>0$ is a parameter playing the role of inverse temperature
and controlling the amplitude of a noise added to the dynamics. Note
the specific scale on which the regularizing and the noise terms act
in~\eqref{eq:34fT}: they are higher order perturbations. We
comment on the choice of this scaling in Remark~\ref{rem:noisescaling}
below. The SDEs~\eqref{eq:34fT} are overdamped Langevin equations
associated with the energy:
\begin{equation}
  \label{eq:66fT}
  \begin{aligned}
    E_\beta(\thetab_1,\ldots, \thetab_n) &= nC_f -\sum_{i=1}^n  F(\thetab_i)
    +\frac1{2n} \sum_{i,j=1}^n  K(\thetab_i,\thetab_j)\\
    &  - (\beta n)^{-1} \sum_{i=1}^n \log \rho(\thetab_i),
  \end{aligned}
\end{equation}
This energy is~\eqref{eq:66} plus a regularizing term (the one
involving $ - \log \rho_0$). Under Assumption~\ref{as:meas} this term guarantees that, for any
$\beta>0$, the following integral is finite
\begin{equation}
  \label{eq:67}
  Z_n = \int_{D^n} e^{-n \beta E_\beta(\thetab_1,\ldots,\thetab_n)} d\thetab_1 \cdots d\thetab_n<\infty
\end{equation}
which in turns implies that
\begin{equation}
  \label{eq:35}
  Z^{-1}_n \exp\left(-n \beta  E_\beta (\thetab_1,\ldots, \thetab_n) \right)
\end{equation}
is a normalized probability density on $D^n$. As a result,
the solutions to~\eqref{eq:34} are ergodic with respect to the
equilibrium distribution with density~\eqref{eq:35} for any
$\beta>0$.

\subsection{Dean's equation}
\label{sec:deanfT}

Let $\chi:D\to\RR$ be a test function. Applying It\^o's
formula to $n^{-1}\sum_{i=1}^n \chi(\thetab_i(t)) = \int_D  \chi(\thetab)\mu_t^{(n)}(d\thetab)$ and using~\eqref{eq:34fT} gives
\begin{equation}
  \label{eq:42fT}
  \begin{aligned}
    & d \int_D \chi(\thetab) \mu_t^{(n)} (d\thetab)\\
    & = \frac1n \sum_{i=1}^n
     \nabla \chi(\thetab_i(t))\cdot d\thetab_i(t)  + \beta^{-1} \sum_{i=1}^n  
     \Delta\chi(\thetab_i(t))dt\\
    & = \int_D
     \nabla \chi(\thetab)\cdot \left( \nabla F(\thetab) \mu_t^{(n)}(\thetab) -
        \int_{D}\nabla K(\thetab,\thetab') \mu_t^{(n)}(d\thetab') \mu_t^{(n)}(d\thetab) 
        \right)dt\\
        & + (\beta
        n)^{-1} \int_D
     \nabla \chi(\thetab)\cdot  \left(\nabla \log \rho_0(\thetab) \, \mu_t^{(n)}(d\thetab) 
        \right)dt\\
    &  + (\beta n)^{-1} \int_D
     \Delta \chi(\thetab) \mu_t^{(n)}(d\thetab) dt \\
    &+ \sqrt{2} (\beta n^{3})^{-1/2} \sum_{i=1}^n  \nabla\chi(\thetab_i(t)) \cdot d\Wb_i(t)
  \end{aligned}
\end{equation}
The drift terms in this equation are closed in terms of $\mu^{(n)}_t$; the noise term has a 
quadratic variation given by:
\begin{equation}
  \label{eq:46fT}
  \begin{aligned}
    & \left\langle  (\beta n^{3})^{-1/2} \sum_{i=1}^n  \nabla\chi(\thetab_i(t)) \cdot d\Wb_i(t),  (\beta n^{3})^{-1/2} \sum_{i=1}^n  \nabla\chi(\thetab_i(t)) \cdot d\Wb_i(t)\right\rangle\\
    & =   \beta^{-1} n^{-3} \sum_{i=1}^n |\nabla\chi(\thetab_i(t))|^2 dt
      \\
    & =   \beta^{-1} n^{-2}\int_D  |\nabla\chi(\thetab)|^2 \mu^{(n)}_t(d\thetab) dt
  \end{aligned}
\end{equation}
As a result, \eqref{eq:42fT} is sometimes written formally as the stochastic partial differential equation (SPDE)
\begin{equation}
  \label{eq:38fT}
  \begin{aligned}
    \partial_t \mu_t^{(n)} & = \nabla \cdot\left( \nabla F \mu_t^{(n)} +
        \int_{D}\nabla K(\thetab,\thetab') \mu^{(n)}(d\thetab') \mu_t^{(n)} 
        \right)\\
        & - (\beta
        n)^{-1} \nabla \cdot\left(\nabla \log \rho_0 \, \mu_t^{(n)} 
        \right)\\
    &  + (\beta n)^{-1} \Delta \mu_t^{(n)} + \sqrt{2}\beta^{-1/2} n^{-1}
    \, \nabla \cdot\big( \sqrt{\mu_t^{(n)}}\,
      \dot\etab_t\big) 
  \end{aligned}
\end{equation}
where $\dot\etab_t=\dot\etab_t(\theta)$ is a  spatio-temporal
white-noise so that the quadratic variation of the noise term in~\eqref{eq:38fT} is formally given by~\eqref{eq:46fT}. This equation is referred to as Dean's equation. It is difficult to give \eqref{eq:38} a precise meaning because it is not clear how to interpret the noise term. It
remains useful to analyze the properties of $\mu_t^{(n)}$ as $n\to\infty$,
however, which is what we will do next.

\begin{remark}
  \label{rem:noisescaling}
  We could also consider situations where in~\eqref{eq:34fT} $n^{-1}$
  is replaced by $n^{-\alpha}$ with $\alpha \in [0,1)$. The case
  $\alpha=0$ is treated in\cite{Mei:2018}: with this scaling, the
  diffusive and regularizing terms in~\eqref{eq:38fT} are replaced by
  $$\beta^{-1} \Delta \mu_t^{(n)} 
  -\beta^{-1}\nabla\cdot \left( \nabla \log \rho_0 \, \mu_t^{(n)} \right),$$ and
  the noise terms by
  $$\sqrt{2}(\beta n)^{-1/2} \, \nabla \cdot\big( \sqrt{\mu_t^{(n)}}\,
    \dot\etab_t\big).$$
    This means that these
  diffusive and regularizing terms affect the mean field limit
  equation for $\mu_t$, whereas the noise terms remain higher
  order. In particular, in that case one can prove that
  $\mu^{(n)}_t\rightharpoonup \mu_t$ with $\mu_t$ that converges
  to a unique fixed point $\mu_\beta$ such that $\mu_{\beta}>0$
  a.e. in $D$ but for which
  $\int_{D} \varphi(\cdot,\thetab) \mu_{\beta}(d\thetab) 
  \not = f$ (there is a correction proportional to $\beta^{-1}$).
  When $\alpha \in (0,1)$, the diffusive and regularizing terms
  in~\eqref{eq:38fT} are replaced by
  $$\beta^{-1} n^{-\alpha} \Delta \mu_t^{(n)} -\beta^{-1} n^{-\alpha}\nabla\cdot
  \left( \nabla \log \rho_0\,\mu_t^{(n)} \right),$$ and the noise terms by
  $$\sqrt{2}(\beta n^{1+\alpha})^{-1/2} \, \nabla \cdot\big(
    \sqrt{\mu^{(n)}}\, \dot\etab_t\big).$$
    This means that none of these terms affect the
  mean field limit equation, but at next order, $O(n^{-\alpha})$, the
  diffusive and regularizing terms dominate whereas the noise terms
  remain higher order. In the case when $\alpha=1$, on which we focus
  here, the diffusive, regularizing, and noise terms are
  perturbations on the $O(n^{-1})$ same scale, the same scale
  as the errors introduced by discretization effects (finite $n$)
  also present in GD.
\end{remark}

\subsection{Multiple-scale expansion}

The advantage of adding noise terms in~\eqref{eq:34fT} is that it
guarantees ergodicity of the solution to these SDEs with respect to
the equilibrium distribution with
density~\eqref{eq:35}. Correspondingly, we focus on analyzing the
long-time ergodicity properties of the empirical distribution
satisfying~\eqref{eq:38fT}. On long timescales, the memory of
the initial conditions is lost, and we can directly pick the right
scaling to analyze the fluctuations of $\mu^{(n)}_t$ around its limit
$\mu_t$: as discussed in Remark~\ref{rem:noisescaling} and
confirmed below, this scale is $O(n^{-1})$, consistent with what we
reach at long times with GD as discussed in Sec.~\ref{sec:weighted}.

We analyze~\eqref{eq:38fT} by formal asymptotic, using a
two-timescale expansion. Consistent with the expected $O(n^{-1})$
scaling of the fluctuations, we look for a solution of this equation
of the form
\begin{equation}
  \label{eq:4}
  \mu^{(n)}_t = \mu_{t,\tau}+n^{-1} \omega_{t,\tau}+o(n^{-1}), \qquad \tau=t/n.
\end{equation}
We use the rescaled time $\tau=t/n$ to look at the solution
to~\eqref{eq:38fT} on $O(n)$ timescales.  Not only does this fix the
behavior of $\mu_{t,\tau}$ on long timescales but also
guarantees solvability of the equation for $\omega_{t,\tau}$. Treating $t$
and $\tau$ as independent variables, \eqref{eq:4} implies that
\begin{equation}
  \label{eq:6}
  \partial_t \mu^{(n)}_t = \partial_t \mu_{t,\tau} + n^{-1} (\partial_\tau \mu_{t,\tau}  + \partial_t \omega_{t,\tau})
\end{equation}
Inserting~\eqref{eq:4} and~\eqref{eq:6} in~\eqref{eq:38fT} and
collecting terms of the same order in $n^{-1}$, we arrive at the
following two equations at order $O(1)$ and $O(n^{-1})$, respectively
\begin{equation}
  \label{eq:38clt0}
    \partial_t \mu_{t,\tau} = \nabla \cdot\left(
      \nabla V(\thetab,[\mu_{t,\tau}]) \mu_{t,\tau} \right) \\
\end{equation}
and
\begin{equation}
  \label{eq:38clt}
  \begin{aligned}
    \partial_\tau \mu_{t,\tau} + \partial_t \omega_{t,\tau}& = \nabla \cdot\left( 
      \nabla V(\thetab,[ \mu_{t,\tau}]) \omega_{t,\tau} + \nabla F(\thetab,[\omega_{t,\tau}]) \mu_{t,\tau}\right) \\
    & + \beta^{-1} \Delta \mu_{t,\tau} - \beta^{-1} \nabla\cdot
    \left( \nabla
      \log \rho_0\, \mu_{t,\tau}\right)\\
    & + \sqrt{2}\beta^{-1/2} \, \nabla \cdot\left( \sqrt{\mu_{t,\tau}}
      \dot\etab_t\right) 
  \end{aligned}
\end{equation}

\subsection{Law of Large Numbers at finite temperature}

Since~\eqref{eq:38clt0} is identical to~\eqref{eq:39}, the results
we established in Sec.~\ref{sec:lln} still hold at finite
temperature. In particular, Proposition~\ref{th:lln} applies. As we
see below, we can obtain more information about $\mu_t$ by looking
at the evolution of this function on longer timescales, and we will
be able to deduce that $\supp \mu_{t,\tau}=D$. This
guarantees that~\eqref{eq:32} holds, so it can be removed from the
assumptions needed in Proposition~\ref{th:lln}.

\subsection{Global convergence on
  $O(n)$ timescales}

An equation governing the evolution of $\mu_{t,\tau}$ on the rescaled time
$\tau=t/n$ can be derived by time averaging~\eqref{eq:38clt} over
$t$. This equation guarantees the solvability
of~\eqref{eq:38clt}. Since $\mu_{t,\tau} \to \mu_{\tau}$ as
$t\to\infty$, where $\mu_{\tau}$ is a stationary point
of~\eqref{eq:38clt0}, we have
\begin{equation}
  \label{eq:43}
  \lim_{T\to\infty} \frac1T \int_0^T \mu_{t,\tau} \omega_{t,\tau} dt =
  \mu_\tau\bar \omega_\tau     
\end{equation}
where
\begin{equation}
    \bar
  \omega_\tau =: \lim_{T\to\infty} \frac1T \int_0^T \omega_{t,\tau} dt
\end{equation}
in which we  assume that the time-average of $\omega_{t,\tau}$ exists (which we
check \textit{a~posteriori}). Using~\eqref{eq:43} and the fact that
the white-noise terms time-average to zero almost surely, we deduce
that the time-average of~\eqref{eq:38clt} is
\begin{equation}
  \label{eq:53tavgaa}
  \begin{aligned}
    \partial_\tau \mu_\tau & = \nabla \cdot\left( \nabla V(\thetab,[\mu_\tau]) \bar \omega_\tau + \nabla F(\thetab,[\bar \omega_{\tau}])\mu_\tau \right)\\
    & + \beta^{-1} \Delta \mu_\tau - \beta^{-1} \nabla\cdot
    \left( \nabla
      \log \rho_0\, \mu_{\tau}\right)
  \end{aligned}
\end{equation}
Because of the presence of the diffusive
term $\beta^{-1} \Delta \mu_\tau$
in~\eqref{eq:53tavgaa}, we can therefore conclude that on the timescales
where this equation holds we must have $\mu_\tau>0$ a.e. on
$D$. This means that~\eqref{eq:32} holds and so
$V(\thetab,[\mu_\tau])=0$ since $\int_D \varphi(\cdot, \thetab) \mu_\tau(d\theta) = f$ by
Proposition~\ref{th:lln}.  As a result \eqref{eq:53tavgaa} reduces to
\begin{equation}
  \label{eq:53tavg}
  \begin{aligned}
    \partial_\tau \mu_\tau & = \nabla \cdot\left( \nabla F(\thetab,[\bar \omega_{\tau}])\mu_\tau \right) + \beta^{-1} \Delta \mu_\tau - \beta^{-1} \nabla\cdot
    \left( \nabla
      \log \rho_0\, \mu_{\tau}\right).
  \end{aligned}
\end{equation}
Since $V(\thetab,[\mu_\tau])=0$ needs to be satisfied, in~\eqref{eq:53tavg} we can
treat the term involving the factor $F(\thetab,[\bar \omega_{\tau}])$ as a Lagrange multiplier
used to enforce this constraint. It is also
easy to see that \eqref{eq:53tavg} is the Wasserstein GD flow on
\begin{equation}
  \label{eq:129}
  \begin{aligned}
    \int_{D}
    \left(\beta^{-1}\log (d\mu/d\mu_0) +  F(\thetab,[\bar \omega_{\tau}]) \right)\mu(d\thetab)
  \end{aligned}
\end{equation}
Since this energy is strictly convex, 
a direct consequence of these observations is that the stable fixed
points of~\eqref{eq:53tavg} are the minimizers of the
energy~\eqref{eq:129} subject to the constraints that $V(\thetab,[\mu_\tau])=0$ and $\mu_\tau\in \mathcal{M}_+(D)$. These fixed
points are reached on a timescale that is large compared the $O(n)$
timescale $\tau=t/n$.

Recalling that $\mu_{t,\tau}$ is the weak limit of $\mu^{(n)}_t$ as
$n\to\infty$ and $V(\thetab,[\mu]) = -F(\theta)+\int_D K(\thetab,\thetab') \mu(d\thetab')$, we can summarize these considerations into:

\begin{proposition}
  \label{th:lemrho0}
  If $\mu^{(n)}_t$ be the empirical distribution defined
  in~\eqref{eq:37} with $\{\thetab_i(t)\}_{i=1}^n$ the solution
  to~\eqref{eq:38fT}. Then given any $b_n>0$ such that
  $b_n/n\to\infty$ as $n\to\infty$, we have
  \begin{equation}
    \label{eq:86llnrho}
    \mu^{(n)}_{b_n} \rightharpoonup \mu^* \qquad
    \text{as \ \ $n\to\infty$}
  \end{equation}
  where $\mu^*$ is the minimizer in $\mathcal{M}_+(D)$ of
  \begin{equation}
    \label{eq:102}
    \beta^{-1} \int_{D} \log
    (d\mu/d\mu_0) \mu(d\thetab)
  \end{equation}
subject to
 \begin{equation}
    \label{eq:102b}
    F(\thetab)=\int_D K(\thetab,\thetab')\mu^*(d\thetab')  \quad
    \text{a.e. \  in \  $D$}
  \end{equation}
\end{proposition}
\noindent
It is  easy to see that the solution to the minimization problem
in Proposition~\ref{th:lemrho0} is such that
  \begin{equation}
    \label{eq:7}
    \int_{D}  \log(d\mu^*/d\mu_0)\mu^*(d\thetab)<\infty
    \qquad\text{and} \qquad \supp \mu^* = D.
\end{equation}
The first condition says that the minimizer exists, which is clear since we can find test distributions $\mu\in \mathcal{M}_+(D)$ such that (i)
  $ \int_{\RR} c \mu(dc,\cdot)= \gamma^* $ where $\gamma^*$
  solves~\eqref{eq:29} (i.e. such that $\mu$ satisfies the constraint
  in~\eqref{eq:102b}), and (ii) $\mu$ has finite entropy with respect to $\mu_0$. One such $\mu$ is 
\begin{equation}
  \label{eq:110}
  \mu(dc,d\zb) = |\gamma^*|^{-1}_{\text{TV}} \left(\delta_{|\gamma^*|_{\text{TV}}}(dc) \gamma^*_+(d\zb) + \delta_{-|\gamma^*|_{\text{TV}}}(dc) \gamma_-^*(d\zb)\right)
\end{equation}
To prove that $\supp \mu^*= D$, suppose
by contradiction that the minimizer is such that $\mu^*=0$ if
$\thetab\in B$ with $\int_{B} \mu_0(d\thetab)>0$. For $s\in [0,1]$,
consider $\mu^s= (1-s)\mu^*+ s \mu_0$. A direct calculation shows that
\begin{equation}
  \label{eq:20}
  \begin{aligned}
    \int_{D} \log (d\mu^s/d\mu_0) \mu^s(d\thetab)& = \int_{B}
  \log (d\mu^*/d\mu_0) \mu^* (d\thetab)\\
  &+ s\log s \int_{B^c}
  \mu_0(d\thetab) +O(s)
  \end{aligned}
\end{equation}
Since $s\log s \int_{B^c}
  \mu_0(d\thetab)<0 $ for
$s\in(0,1)$, \eqref{eq:20} implies that for $s>0$ small enough
\begin{equation}
  \label{eq:20b}
  \int_{D} \log (d\mu^s/d\mu_0) \mu^s(d\thetab) < \int_{B}
  \log (d\mu^*/d\mu_0) \mu^* (d\thetab),
\end{equation}
a contradiction with our  assumption that $\mu^*$ is the minimizer.

Let us also analyze in some more detail  the constrained optimization problem in Proposition~\ref{th:lemrho0} since this will be useful in the next section.
If we denote by $\mu^*$ the minimizer of~\eqref{eq:102}
subject to~\eqref{eq:102b} and by $\lambda^*$ the Lagrange multiplier
used to satisfy the first constraint in~\eqref{eq:102b}, this Lagrange
multiplier is given by
\begin{equation}
  \label{eq:77}
  \lambda^*(\thetab) = \beta^{-1} \frac{\delta }{\delta F(\thetab)}
  \int_{D} \log (d\mu^*/d\mu_0) \mu^*(d\thetab')
\end{equation}
It is easy to see that $\mu^*$ is independent of $\beta$: indeed,
we can drop the factor $\beta^{-1}$ in front of~\eqref{eq:102} without
affecting the minimization problem. This also means that the
dependency of $\lambda^*$ in $\beta$ is explicit: Indeed
from~\eqref{eq:77}
\begin{equation}
  \label{eq:78}
  \lambda^*(\thetab) = \beta^{-1} \delta^*(\thetab)
\end{equation}
where $\delta^*(\thetab)$ is given by
\begin{equation}
  \label{eq:77b}
  \delta^*(\thetab) = \frac{\delta }{\delta F (\thetab)}
  \int_{D} \log (d\mu^*/d\mu_0) \mu^*(d\thetab')
\end{equation}
This factor is independent of $\beta$ since $\mu^*$ is. It will be useful later to work with the function  $\epsilon^*(\xb)$ defined via the equation
\begin{equation}
  \label{eq:122}
  \int_D \varphi(\xb,\thetab) \delta^*(\thetab) d\thetab = \int_D \int_\Omega
  \varphi(\xb,\thetab) \varphi(\xb',\thetab)
  \epsilon^*(\xb') \nu(d\xb') d\thetab
\end{equation}
This is the Euler-Lagrange equation for the minimizer of
\begin{equation}
  \label{eq:128}
  \tfrac 12 \int_D \left|\delta^*(\thetab) - \int_\Omega \epsilon(\xb)
    \varphi(\xb,\thetab) \nu(d\xb)\right|^2 d\thetab
\end{equation}
over $\epsilon$. Therefore,~\eqref{eq:122} is also the equation for
the least square solution of
\begin{equation}
  \label{eq:22}
  \delta^*(\thetab) = \int_\Omega \epsilon^*(\xb)\varphi(\xb,\thetab) \nu(d\xb)
\end{equation}
and  such a least square solution exists for a
modification of $\delta^*(\thetab)$ which is arbitrarily close to it in
$L^2(D)$: any such solution for a modification of $\delta^*(\thetab)$ that
is $O(n^{-1})$ away from it is good enough for our purpose since the
discrepancy can be absorbed in higher order terms in our expansion in
$n^{-1}$. This solution is also unique by Assumption~\ref{th:as1} and
it can be expressed as
\begin{equation}
  \label{eq:77c}
  \epsilon^*(\xb) = D_{f(\xb)} \int_{D}
  \log (d\mu^*/d\mu_0) \mu^*(d\thetab)
\end{equation}
where $\mu^*$ is viewed as a functional of $f(\xb)$ by using $F(\thetab) = \int_\Omega f(\xb) \varphi(\xb,\thetab) \nu(d\xb)$,
and $D_{f(\xb)}$ denotes the gradient with respect to $f(\xb)$ in the
$L^2(\Omega,\nu)$-norm defined in \eqref{eq:104}. The equality
\eqref{eq:77c} follows from~\eqref{eq:77b} and the fact that
$D_{f(\xb)} F(\thetab) = \varphi(\xb,\thetab)$.

\begin{remark}
  \label{rem:comparison}
  Compared to the case treated in~\cite{Mei:2018} where the noise and
  regularizing terms in~\eqref{eq:34fT} are scaled as $\beta^{-1}$
  (high temperature) rather than $(\beta n)^{-1}$ (low temperature),
  we see that we can also conclude that $\mu_t$ converges as $t\to\infty$ to a
  distribution $\mu^*$ with $\supp \mu^*=D$;
  however, the fixed point $\mu^*$ we obtain satisfies
  $\int_{D}  \varphi(\cdot,\thetab) \mu^* (d\thetab)= f$,
  whereas the one obtained at high temperature introduces a correction
  proportional to $\beta^{-1}$ in this relation. The price we pay by
  working at low temperature is that convergence in time may be slower
  if the initial condition $\mu_0  = \mu_{\text{in}}$ is such
  that~\eqref{eq:32} is not satisfied by the GD flow without noise:
  specifically, this convergence should occur on timescales that are
  intermediate between $O(1)$ and $O(n)$.
\end{remark}

\subsection{Central Limit Theorem at finite temperature}
\label{sec:CLTfT}

Now that we have determined the behavior of
$\lim_{n\to\infty} \mu^{(n)}_t = \mu_t$ at all times, we can stop
distinguishing $\tau$ from $t$, and focus on $\omega_t$. We
already know that~\eqref{eq:102b} constrain the average value of
$\omega_t$ on long timescales, but we would also like to quantify this
average value beyond what~\eqref{eq:102b} implies, and also analyze the fluctuations around
this average. To this end, let us use~\eqref{eq:53tavg} in~\eqref{eq:38clt} and
look at the resulting equation on timescales where
$\mu_t$ has converged to $\mu^*$, the minimizer specified in
Proposition~\ref{th:lemrho0}, so that $V(\thetab,[\mu^*])=0$  and $\lambda$ has converged to $\lambda^*=\beta^{-1} \delta^*$. This can be achieved
by considering~\eqref{eq:38clt} with initial condition at $t=T$ and
pushing back $T\to-\infty$. The resulting equation is
\begin{equation}
  \label{eq:38cltssb}
  \begin{aligned}
    \partial_t \omega_t & = \nabla \cdot\left( - \beta^{-1} \nabla
      \delta^* \mu^* + \int_{D}\nabla
      K(\thetab,\thetab') \omega_t(d\thetab')\mu^*
      \right) \\
    & + \sqrt{2}\beta^{-1/2} \, \nabla \cdot\left( \sqrt{\mu^*}
      \dot\etab_t\right)
  \end{aligned}
\end{equation}
Even though we derived it formally, the SPDE~\eqref{eq:94} can be
given a precise meaning: since its drift is linear in $\omega_t$ and its
noise is additive (recall that $\mu^*$ is a given, non-random,
distribution), \eqref{eq:94} defines $\omega_t$ as a Gaussian process. This
also means that
\begin{equation}
  g_t = \int_{D}
  \varphi(\cdot,d\thetab) \omega_t(d\thetab)
  \label{eq:63fT}
\end{equation}
is a Gaussian process. This is an important quantity since gives the
error on $f$ made in $f^{(n)}_t$ at order $O(n^{-1})$:
\begin{equation}
  \label{eq:100fT}
  f^{(n)}_t =  f + n^{-1} g_t + o(n^{-1})
\end{equation}

Let us derive a closed equation for $g_t$
from~\eqref{eq:38cltssb}. To this end, notice first that we can use
\begin{equation}
  \label{eq:104b}
  \int_D K(\thetab,\thetab') \omega_t(d\thetab') =
  \int_{\Omega} \varphi(\xb,\thetab) g_t(\xb) \nu(d\xb)
\end{equation}
to express the integral terms in~\eqref{eq:38cltssb} in terms of
$g_t$. By taking the time derivative of~\eqref{eq:63fT} and
using~\eqref{eq:38cltssb} together with~\eqref{eq:104b}
and~\eqref{eq:122} we derive:
\begin{equation}
  \label{eq:99}
  \begin{aligned}
    \partial_t g_t &= \int_{D}
    \varphi(\cdot,\thetab) \partial_t\omega_t(d\thetab)\\
    % & = \int_{D} \nabla_{\thetab}\varphi(\cdot,\thetab)  \cdot\left(
    %   \left(-\beta^{-1} \nabla \delta^*(\thetab) + \int_{\Omega}
    %     \nabla_{\thetab} \varphi(\xb',\thetab)g_t(\xb') \nu(d\xb')\right)
    %   \mu^*(d\thetab)\right)\\
    % &\quad - \sqrt{2} \beta^{-1/2} \, \int_{D}
    % \nabla_{\thetab}\varphi(\cdot,\thetab)\cdot \sqrt{\mu^*}
    %     \dot\etab\\
    & = \int_{D} \nabla_{\thetab}\varphi(\cdot,\thetab)  \cdot
      \int_{\Omega}
        \nabla_{\thetab} \varphi(\xb',\thetab)\left( g_t(\xb')- \beta^{-1} \epsilon^*(\xb')\right) \nu(d\xb')
      \mu^*(d\thetab)\\
    &\quad - \sqrt{2} \beta^{-1/2} \, \int_{D}
    \nabla_{\thetab}\varphi(\cdot,\thetab)\cdot \sqrt{\mu^*}
        \dot\etab_t\\
    & = 
      \int_{\Omega}
        M([\mu^*],\xb,\xb') \left( g_t(\xb')- \beta^{-1} \epsilon^*(\xb')\right) \nu(d\xb')\\
    &\quad - \sqrt{2} \beta^{-1/2} \, \int_{D}
    \nabla_{\thetab}\varphi(\cdot,\thetab)\cdot \sqrt{\mu^*}
        \dot\etab_t
  \end{aligned}
\end{equation}
where $M([\mu],\xb,\xb')$ is the kernel defined
in~\eqref{eq:96}.
Since the quadratic variation of the noise
term in this equation is 
\begin{equation}
  \label{eq:41}
  2\beta^{-1} M([\mu^*],\xb,\xb') dt
\end{equation}
in law it is equivalent to
\begin{equation}
  \label{eq:94}
  \begin{aligned}
    \partial_t g_t & = - \int_\Omega M([\mu^*],\xb,\xb')
    \left(g_t(\xb') -\beta^{-1} \epsilon^*(\xb)\right) \nu(d\xb')\\
    &\quad+ \sqrt{2}\beta^{-1/2} \int_\Omega
    \sigma([\mu^*],\xb,\xb') \dot\eta_t(\xb') d\xb'
  \end{aligned}
\end{equation}
where $\dot \eta_t(\xb)$ is a spatio-temporal white-noise, and
$\sigma([\mu^*],\xb,\xb')$ is such that
\begin{equation}
  \label{eq:95}
  \int_\Omega \sigma([\mu^*],\xb,\xb'')
  \sigma([\mu^*],\xb',\xb'') d\xb''
  = M([\mu^*],\xb,\xb')
\end{equation}
Note that this decomposition exists since $\mu^*\in \mathcal{M}_+(D)$ with $\supp \mu^* = D$ and hence $M([\mu^*],\xb,\xb')$ is
positive-definite. The asymptotic mean and variance of $g_t$ can be
readily deduced from~\eqref{eq:94} by noting that this Ornstein-Uhlenbeck equation is in detailed-balance with respect to the Gibbs distribution associated with the energy 
\begin{equation}
    \frac12 \int_\Omega\left|g_t(\xb') -\beta^{-1} \epsilon^*(\xb)\right|^2 \nu(d\xb).
\end{equation} 
We can state this as
\begin{proposition}[CLT at finite temperature]
  \label{th:cltfT}Let $f^{(n)}_t$ be given by~\eqref{eq:68}
  with $\{\thetab_i(t)\}_{i=1}^n$ solution
  to~\eqref{eq:38fT} with initial conditions specified at $t=T$. Then
  \begin{equation}
    \label{eq:86cltfT}
    \lim_{T\to-\infty} \lim_{n\to\infty} n\big(f^{(n)}_t -
      f\big)  = g_t \qquad \text{in law}
  \end{equation}
  where $g_t$ is the stationary Gaussian process specified by~\eqref{eq:94}
  and whose mean an covariance satisfy: for any test function
  $\chi:\Omega\to \RR$
  \begin{equation}
    \label{eq:124}
    \begin{aligned}
      & \EE \int_\Omega \chi(\xb) g_t(\xb)
      \nu(d\xb)
      = \beta^{-1} \int_\Omega \chi(\xb) \epsilon^*(\xb) \nu(d\xb)\\
      & \EE \left(\int_\Omega \chi(\xb)
        \left(g_t(\xb)- \beta^{-1} \epsilon^*(\xb)\right)
        \nu(d\xb)\right)^2 = \beta^{-1} \int_\Omega
      |\chi(\xb)|^2\nu(d\xb)
    \end{aligned}
  \end{equation}
where $\epsilon^*$ is given by~\eqref{eq:77c}
\end{proposition}

\noindent
Notice that if we quench the result in~\eqref{eq:124} (i.e. send
$\beta\to\infty$), we arrive at the conclusion that $g_t \to0$ as
$t\to\infty$ in that case. This is consistent with what happens at
zero-temperature, in the limit as $\xi\to1$, see
Proposition~\ref{th:cltgvlt}. % Notice also that Proposition~\ref{th:cltfT}

\bibliographystyle{alpha}
\bibliography{mlrefs}

\end{document}